\documentclass[11pt]{article}
\usepackage[margin=1in]{geometry}
\usepackage[T1]{fontenc}
\usepackage[utf8]{inputenc}
\usepackage{lmodern}
\usepackage{microtype}
\usepackage{graphicx}
\usepackage{booktabs}
\usepackage{array}
\usepackage{longtable}
\usepackage{tabularx}
\usepackage{enumitem}
\usepackage{amsmath}
\usepackage{amssymb}
\usepackage{natbib}
\usepackage{url}
\usepackage{placeins}
\usepackage{listings}
\usepackage{ragged2e}
\usepackage[colorlinks=true,linkcolor=blue,citecolor=blue,urlcolor=blue]{hyperref}
\graphicspath{{figures/}}
\title{Autonomous Event-Driven Multi-Agent Orchestration for Enterprise AI at Scale}
\author{
Harsh Rao Dhanyamraju\\
Data Scientist, SAP SE
\and
Leonidas Raghav\\
ML Engineer, SAP SE
\and
Aaron Lee\\
ML Engineer, SAP SE
}
\date{}
\hypersetup{
  pdftitle={Autonomous Event-Driven Multi-Agent Orchestration for Enterprise AI at Scale},
  pdfauthor={Harsh Rao Dhanyamraju, Leonidas Raghav, Aaron Lee}
}
\begin{document}
\raggedbottom
\setlength{\textfloatsep}{8pt plus 2pt minus 2pt}
\setlist[itemize]{leftmargin=*,itemsep=0.15em,topsep=0.25em}
\maketitle
\begin{center}
\small\texttt{\{harsh.rao.dhanyamraju,leonidas.raghav,aaron.lee01\}@sap.com}
\end{center}
\begin{abstract}
Enterprise agent language models must coordinate specialist agents over continuous business-event streams, yet most multi-agent evaluations assume discrete request-response workflows. We present a controlled simulation based on a sanitized enterprise research snapshot: 208 scenarios, 393 events, and 1,051 expected agent calls across Persona (<10 agents), Department (20--80), and Enterprise (200) registries. We compare DAG Plan \& Execute with ReAct and evaluate a minimal Task Manager for priority inference, related-event merging, and preemption. Both systems perform well on the Persona cohort but score lower on the Enterprise cohort, especially for Simple scenarios; a larger, more confusable discovery space may contribute. The tested DAG Plan \& Execute implementation generally maintains higher agent-call precision, whereas ReAct shows stronger recall and Expected Answer Completeness in several Enterprise conditions. The Task Manager reduces high-priority queue latency by 14--75\% and improves related-event Expected Answer Completeness by over 20 percentage points at Enterprise scale. These controlled-simulation results characterize the tested systems but do not establish production readiness.
\end{abstract}

\section{Introduction}
\subsection{OpenClaw \& Multi-Agent Systems}

AI agents extend LLMs beyond text generation by observing an environment, reasoning over goals, and invoking tools or APIs to complete tasks. OpenClaw's rapid adoption in 2026 showed strong demand for this model: its continuous reasoning loop and standardized tool integration let agents browse, edit files, query databases, and coordinate services across real operating environments \citep{steinberger2026openclaw}. That same integration creates the central risk. By combining untrusted inputs, autonomous action, extensibility, and privileged system access in one loop, OpenClaw-style systems can be insecure by design \citep{li2026defensible}; unconstrained deployments such as Moltbook demonstrate how weak guardrails can produce safety incidents and unregulated behavior at scale \citep{manik2026agentonly}. For enterprise settings, autonomy therefore cannot be treated as model capability alone: it requires structural guardrails, permission boundaries, validation logic, and deterministic fallback paths that keep agent actions within operational and business constraints.

Subsequent frameworks such as NanoClaw \citep{qwibit2026nanoclaw} and IronClaw \citep{nearai2026ironclaw} respond with container sandboxing, encrypted local state, and layered controls, underscoring that agent usefulness depends on constraining the same system access that makes agents powerful.

Beyond security, the limits of single agents have pushed the field toward multi-agent architectures \citep{adimulam2026orchestration}: context and reasoning constraints favor specialized modular agents, while MCP \citep{mcp2026intro} and A2A \citep{surapaneni2025a2a,a2a2026docs} provide interoperability and distributed collectives can be more economical than monolithic agents \citep{dang2025multiagent}.

Prior work already provides orchestration mechanisms, but mostly for discrete tasks: Deep Agent uses hierarchical task DAGs and planner-executor decomposition \citep{zhang2025autonomous}, DynTaskMAS adds dynamic task graphs and asynchronous parallelism \citep{chen2025dyntaskmas}, Anthropic's research system demonstrates orchestrator-worker parallel subagents \citep{anthropic2025multiagent}, and Microsoft codifies coordination topologies such as sequential, concurrent, group chat, handoff, and Magentic patterns \citep{microsoft2026patterns}. Production frameworks including LangGraph \citep{langchain2024langgraph}, AutoGen \citep{wu2024autogen}, IBM Watsonx Orchestrate \citep{ibm2024watsonx}, Crew.ai \citep{moura2024crewai}, PydanticAI \citep{pydantic2024ai}, and Google's Agent Development Kit \citep{google2025adk} further supply graph, conversation, automation, role-based, and agent-development infrastructure. Together, these systems establish that multi-agent coordination is practical, but they do not evaluate continuous enterprise event streams, scale-dependent agent discovery, or task prioritization and merging.

For enterprise deployment, these orchestration mechanisms must be coupled with deterministic guardrails that validate outputs, enforce business rules, and bound agent autonomy.

Yet despite these advances, a critical gap remains: existing systems are designed for request-response interactions, not continuous event-driven operation.

\subsection{The Enterprise Orchestration Challenge}

\begin{figure}[!tbp]
\centering
\includegraphics[alt={Conceptual progression from reactive chatbots to autonomous digital workers that monitor enterprise events, coordinate specialist agents, and present resolutions.},width=0.70\linewidth]{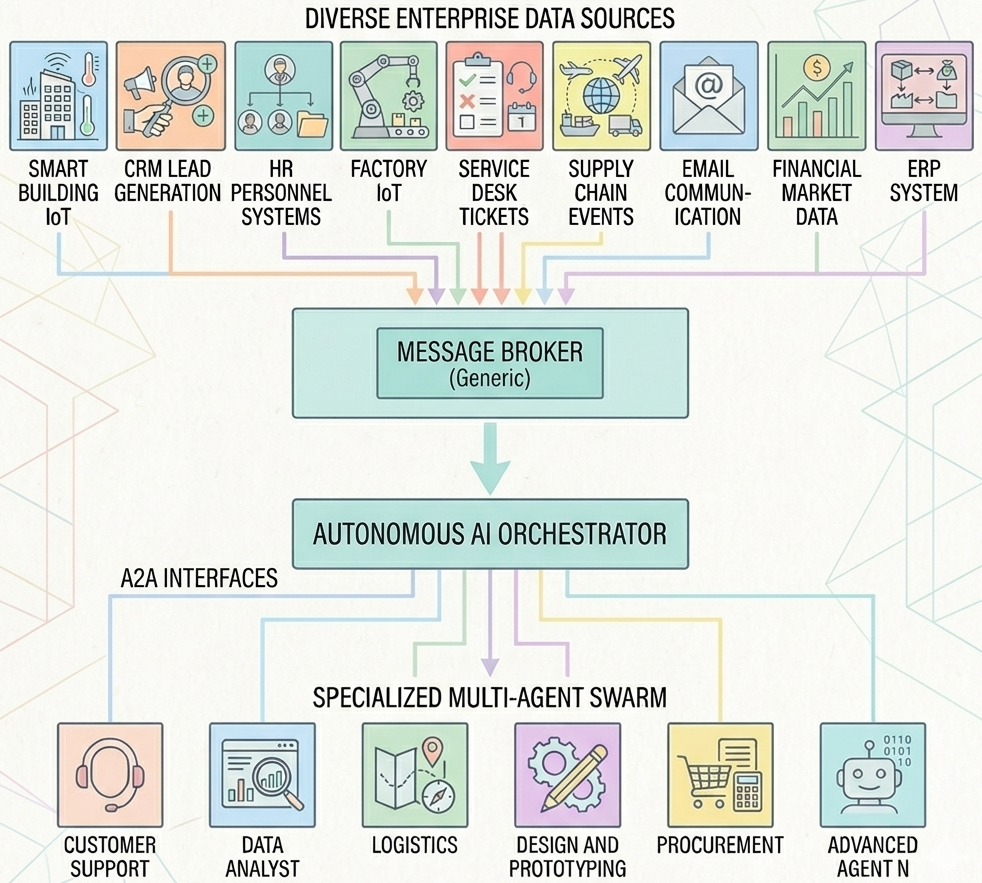}
\caption{Enterprise event-driven orchestration vision.}\label{fig:vision}
\end{figure}

A central goal of enterprise AI is not merely to answer questions when asked, but to monitor, triage, and respond to events as they occur while preserving operational guardrails. SAP's public Autonomous Enterprise direction similarly emphasizes governed coordination of assistants and more than 200 specialized agents across end-to-end business processes \citep{sap2026autonomousenterprise}. Consider an AI system that observes a production error in a factory, recognizes its severity, investigates the root cause by coordinating multiple specialist agents, and presents a resolution before a human operator even notices the alert. Figure~\ref{fig:vision} illustrates this shift from reactive chatbots to autonomous digital workers operating continuously alongside human teams.

Recent benchmarks such as MCP-Bench \citep{wang2026mcpbench}, Tool Decathlon \citep{li2026tooldecathlon}, and MCPMark \citep{wu2026mcpmark} show that planning and coordination, not language understanding alone, differentiate agent performance, and that agents struggle to maintain coherent state across long-horizon tasks. However, these benchmarks focus on single-agent tool use. They do not evaluate multi-agent orchestration at scale, continuous event-driven operation, or task prioritization and merging.

Multi-agent orchestration at enterprise scale remains underexplored. We compare structured DAG Plan \& Execute against ReAct across three organizational scales: Persona (<10 agents), Department (20--80 agents), and Enterprise (200 agents), and four task types: Simple, Parallel, Complex, and Failure.

Existing multi-agent systems assume discrete request-response workflows, while enterprise environments produce asynchronous event streams: alerts, user requests, scheduled jobs, and follow-up signals with different urgency. Related events should merge into a unified task rather than fragment into duplicate work, while critical events should preempt routine processing. We introduce a minimal Task Manager integration layer that infers priority, merges related events, enables preemption, and supports steering by absorbing new events into an active plan. The Task Manager is not intended as a novel scheduling algorithm; it is a controlled mechanism for measuring event-stream capabilities that request-response orchestrators usually omit.

This continuous-event setting also motivates SAP's framing of the Autonomous Enterprise. SAP describes the Autonomous Enterprise as a model in which humans, AI assistants, and agents work together across critical business workflows, supported by Joule Work, SAP Autonomous Suite, and SAP Business AI Platform \citep{sap2026autonomousenterprise,sapSapphireInnovationGuide2026}. SAP further characterizes SAP Business AI Platform as an enterprise AI foundation that combines AI, data, process context, and governance, with components such as SAP Knowledge Graph and Joule Studio intended to ground agents in business semantics and govern their deployment \citep{sapBusinessAIPlatform2026}. For this paper, that framing becomes a concrete research question: how should enterprise AI systems coordinate many specialist agents over continuous business events while preserving process context, user permissions, and operational guardrails? Realizing this goal requires more than capable individual agents. An enterprise orchestrator must select relevant agents from large registries, schedule work across concurrent event streams, preserve user-specific privilege boundaries, merge related events without duplicating work, and route actions through domain- and workflow-specific guardrails. This paper studies that orchestration layer directly: we evaluate how competing agent-control architectures behave as registry scale grows, and whether a Task Manager can support continuous event-driven operation through backlog management, event merging, and preemption.

\section{Test Systems}
\begin{figure}[!t]
\centering
\includegraphics[alt={Task Manager integration points for DAG Plan and Execute and ReAct, including backlog inspection, event merging, checkpoints, and preemption.},width=0.987\linewidth]{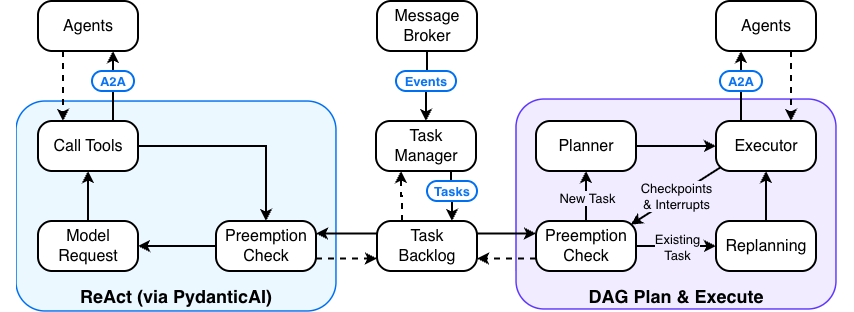}
\caption{Task Manager integration with orchestration architectures.}\label{fig:task-manager-integration}
\end{figure}
\subsection{DAG Plan \& Execute}

DAG Plan \& Execute is a structured orchestration approach where a Planner generates an execution graph upfront, an Executor dispatches work to agents, and a Replanner adapts the plan when conditions change. This architecture separates planning from execution, enabling parallel agent invocation and systematic failure recovery.

The Planner transforms task descriptions into directed acyclic graphs (DAGs). Agent discovery operates in two phases, inspired by Anthropic's Skills system \citep{anthropic2025skills}: first, lightweight summaries of all registered agents are retrieved, allowing the model to select relevant agents without loading full context; second, complete agent cards are loaded only for selected agents. This prevents the Planner from reasoning over hundreds of complete agent descriptions at once while still giving it enough skill-level detail for the agents it actually selects.

The Planner then generates a typed execution graph. Agent nodes represent calls to specialist A2A agents, while combine nodes synthesize intermediate results without making a new external agent call. Edges encode both ordering constraints and context-passing instructions: a downstream node can depend on identifiers, summaries, or partial findings produced upstream. Before execution begins, the DAG is validated for acyclic structure, valid node references, reachable roots, and dependency consistency. This validation step is important because a malformed plan would otherwise fail only after partial execution.

The Executor dispatches nodes to agents via the A2A protocol, an asynchronous pattern where the orchestrator submits tasks and polls for completion. Before each invocation, a lightweight LLM synthesizes the agent prompt from the task context and prior results--since planning is complete, these calls require minimal reasoning. Nodes execute in parallel batches: any node whose prerequisites have completed can run concurrently with other ready nodes. After each response, the LLM evaluates the output: CONTINUE if satisfactory, RETRY with an amended message if incomplete, or INTERRUPT if failure requires plan modification.

Execution is therefore not a single monolithic plan run. The Executor advances through the DAG in batches, records completed nodes and accumulated results in OrchestrationState, and maintains a quantum counter for completed agent calls. A deterministic check monitors this quantum budget; when exhausted, execution pauses at a CHECKPOINT. Figure~\ref{fig:task-manager-integration} shows where this checkpoint boundary connects DAG Plan \& Execute to the Task Manager: before replanning begins, the Task Manager can inspect the backlog, inject merged event context, or preempt the active task for higher-priority work.

The Replanner handles dynamic adaptation, triggered by CHECKPOINT, INTERRUPT, or RESUME (task returning from suspension). A two-level decision process determines: (1) whether replanning is needed, and (2) whether different agents are required. When current agents suffice, the system performs a DAG edit preserving completed work; when new agents are needed, a full replan runs agent discovery again with previous context included.

The Agent Registry implements two-tier discovery (summaries for selection, full cards on-demand). OrchestrationState maintains shared context: current DAG, completed nodes, accumulated results, and quantum counter.

Further architectural details and design tradeoffs are discussed in Appendix~\ref{app:dag-architecture}, and the full DAG planner prompts appear in Appendix~\ref{app:dag-prompts}.

\subsection{ReAct}

ReAct (Reasoning and Acting) \citep{yao2023react} is the default orchestration architecture for LLM-based agents, where a single model interleaves reasoning and action in a continuous loop without upfront planning. Our implementation uses the PydanticAI framework \citep{pydantic2024ai}, which supports parallel tool calling. The agent receives the task and conversation history, emits tool calls--including A2A invocations to specialist agents--observes results, and continues until the task is complete. Native parallel tool execution allows independent agent calls (via A2A) to proceed concurrently, partially mitigating the latency disadvantage of reactive approaches.

In the event-driven setting, ReAct remains a single reasoning loop but exposes deterministic stoppage points to the Task Manager. As shown in Figure~\ref{fig:task-manager-integration}, the Task Manager can inspect the backlog after an agent call returns and before the next reasoning step begins, allowing a higher-priority task to preempt the current one or merged event context to be injected into the next step.

The full ReAct system prompt is provided in Appendix~\ref{app:react-prompt}.

\subsection{Task Manager}

The Task Manager serves as the entry point for all work entering the orchestration system. It is a minimal event-stream integration layer, analogous to a process scheduler in an operating system kernel, used here to make continuous workload behavior measurable. Events from monitoring systems, user interactions, scheduled triggers, or peer assistants arrive via a message broker which is an asynchronous publish-subscribe layer that decouples event sources from the orchestration logic.

The event to task conversion leverages LLM-powered decision making. Events arrive without priority information; when an event arrives, the model decides whether to create a new task or merge it into an existing one, and assigns the appropriate priority level based on event content and context. LLM decisions are validated against configuration constraints; if validation fails, the system re-prompts with error feedback or falls back to safe defaults, ensuring no event is dropped. This component is intentionally simple: the evaluation asks whether priority, merging, and preemption change orchestration behavior, not whether this design is an optimal enterprise scheduler.

Tasks are placed in a priority-ordered backlog. The priority system defines three levels (LOW, MEDIUM, HIGH) with a scoring function incorporating base priority, task age, and progress toward completion--preventing starvation while incentivizing completion of in-progress work. The orchestrator pulls the highest-scoring task from the backlog for execution. Tasks transition through QUEUED (in backlog), ACTIVE (being orchestrated), and COMPLETED states.

The Task Manager integrates with both orchestration architectures at defined stoppage points where preemption or task updates (from merged events) can occur. Figure~\ref{fig:task-manager-integration} shows this integration for both ReAct and DAG Plan \& Execute. For DAG Plan \& Execute, this happens at checkpoints--just before replanning begins--where the system deterministically checks for higher-priority tasks or pending event merges. For ReAct, stoppage points occur after an agent call returns but before the next LLM reasoning step, leveraging PydanticAI's \texttt{agent.iter()} \citep{pydantic2026agentiter} to pause between iterations. At either stoppage point, the system can preempt the current task for higher-priority work or inject merged event context into the active task.

Further details on event processing and priority scoring are provided in Appendix~\ref{app:task-manager-scoring}.

\section{Testing Methodology}
\subsection{Mock Agents}

Testing orchestration at enterprise scale with hundreds of production agents would be prohibitively expensive and difficult to control. Hence, the evaluation harness uses a controlled simulation based on a sanitized enterprise research snapshot: mock agents simulate realistic enterprise behavior while providing controlled ground-truth responses. Each mock agent is an LLM-powered service conforming to the A2A protocol, exposing a realistic agent card with description, skill declarations, and capability metadata. When the System Under Test (SUT) routes a request to a mock agent, it makes a single LLM call that receives the agent's identity, the active story's context, and per-agent guidelines specifying expected entities, keywords, and ground-truth response data.

Entity identification serves as a hard filter enforced by the LLM--the agent only releases information if the request properly identifies the relevant entity (equipment ID, order number, customer name). This mirrors real enterprise behavior and enforces prerequisite dependencies, since entity identifiers are often obtained from previous agent calls. Once identification passes, the agent returns only the specific information requested, even if its guidelines contain additional data; if the entity is not recognized or the request falls outside scope, the agent returns a polite refusal. This design isolates the orchestration properties under study: agent discovery, dependency handling, prompt synthesis, retry behavior, related-event merging, and preemption. It does not substitute for live production validation, but it allows the benchmark to distinguish orchestration failures from downstream tool or data availability failures. We use \textbf{Expected Answer Completeness} as the response-correctness measure: an LLM judge compares the final synthesized response against expected answer fragments. Figures retain the shorter label \emph{Correctness}. As a recall-like metric, it does not penalize extra hallucinated, contradictory, irrelevant, or unsupported content. Further details on mock-agent behavior are provided in Appendix~\ref{app:mock-agent-design}, and the judge prompt is provided in Appendix~\ref{app:judge-prompt}.

\subsection{Evaluation Protocol}

Each scenario is evaluated by submitting only the natural-language event prompt to the system under test; expected agents, call patterns, and expected answer fragments are withheld from the orchestrator and used only for scoring. The main comparison reports scenario-level scores across DAG Plan \& Execute and ReAct at Persona, Department, and Enterprise registry sizes. The reported confidence intervals are computed over scenario-level results and are intended to quantify variation across benchmark cases rather than repeated stochastic reruns of the same case.

The reported experiments use commercial inference-only LLM APIs. DAG planning and ReAct reasoning use the primary model, while lower-complexity executor checks use the fast model as described in the architecture sections. Token and cost analyses in Appendix~\ref{app:extended-results} report the model split used for those runs: Claude Sonnet 4.6 as the normal model and Claude Haiku 4.5 as the fast model. The core orchestration client uses temperature 0 by default for deterministic planner, synthesis, and evaluator calls; mock-agent calls use the mock-agent client defaults. No model training or fine-tuning is performed.

We score four aspects of orchestration behavior. Agent discovery precision and recall compare actual agent calls against expected agent calls, ignoring duplicate calls to the same agent when an orchestrator splits a request across multiple invocations. Execution-pattern correctness checks whether independent and dependent calls are sequenced appropriately. Entity extraction and mock-agent classification measure whether routed prompts include the identifiers needed to unlock ground-truth responses. Final-response Expected Answer Completeness is judged against expected answer fragments rather than open-ended preference rankings.

\subsection{Benchmark Design Rationale}

The benchmark is designed to isolate scale as an orchestration variable. Persona-level scenarios test whether an architecture works when the relevant specialist set is small and discovery is easy. Department-level scenarios introduce a larger but still thematically related agent pool. Enterprise-level scenarios create the needle-in-a-haystack setting that occurs when an assistant must find a few relevant capabilities among hundreds of registered agents.

The scenario taxonomy separates different failure modes:

\begin{itemize}[leftmargin=*,itemsep=0.1em,topsep=0.2em,parsep=0pt]
\item \textbf{Simple scenarios} primarily test discovery precision: the orchestrator must locate one to three relevant agents without being distracted by semantically adjacent tools.
\item \textbf{Parallel scenarios} test whether independent branches are dispatched concurrently.
\item \textbf{Complex scenarios} test mixed execution structure: they combine dependent steps, where downstream agents often need identifiers produced upstream, with parallel branches that can proceed once the required context is available.
\item \textbf{Failure scenarios} test whether the system detects incomplete or ambiguous responses and recovers without cascading errors.
\item \textbf{Priority and related-event scenarios} test whether the Task Manager can support continuous operation rather than one-shot request handling.
\end{itemize}

This design intentionally uses deterministic expected agents, expected call patterns, and expected answer fragments. The goal is not to benchmark open-ended language quality, but to measure whether an orchestration architecture routes, sequences, retries, merges, and preempts work correctly as the agent registry grows.

\section{Dataset}

The evaluation dataset is a sanitized SAP-derived research snapshot comprising 208 scenarios based on enterprise user stories, workflows, and business processes, generating 393 events and 1,051 total agent calls. Scenarios span three organizational scales--Persona (<10 agents), Department (20--80 agents), and Enterprise (200 agents)--to compare orchestration behavior across differently sized discovery spaces. The mock agent population covers 200 enterprise functions organized into seven business domains (Finance, HR, Procurement, Revenue, Supply Chain, Technology Platform, and Customer Support) with 19 specialized personas. It is a sanitized research representation of enterprise agent use cases assembled at SAP SE, not a production-system inventory.

\textbf{Orchestration scenarios} test structural complexity:

\begin{itemize}[leftmargin=*,itemsep=0.1em,topsep=0.2em,parsep=0pt]
\item \emph{Simple} scenarios (1--3 agents) establish baseline competency.
\item \emph{Parallel} scenarios (4--8 agents) test concurrent execution of independent subproblems.
\item \emph{Complex} scenarios (5--10 agents) combine dependency handling with parallel work, creating multi-stage execution paths where upstream agents provide entity identifiers required by downstream agents while independent branches can proceed concurrently.
\item \emph{Failure} scenarios introduce incomplete or ambiguous responses, requiring the orchestrator to detect missing information, retry with refined requests, or proceed with partial data.
\end{itemize}

\textbf{Event management scenarios} test the Task Manager's continuous operation capabilities:

\begin{itemize}[leftmargin=*,itemsep=0.1em,topsep=0.2em,parsep=0pt]
\item \emph{Priority} scenarios verify correct triage when critical events (equipment failures, safety incidents) arrive during lower-priority work.
\item \emph{Related Events} scenarios require identifying semantic relationships between events pertaining to the same underlying situation, merging them into unified tasks, and leveraging cross-event context--while filtering unrelated distractors.
\end{itemize}

\begin{figure}[!t]
\centering
\includegraphics[alt={Grouped bar charts of mean Expected Answer Completeness by scenario type, orchestrator, and Persona, Department, or Enterprise cohort.},width=\linewidth,trim=7bp 8bp 5bp 6bp,clip]{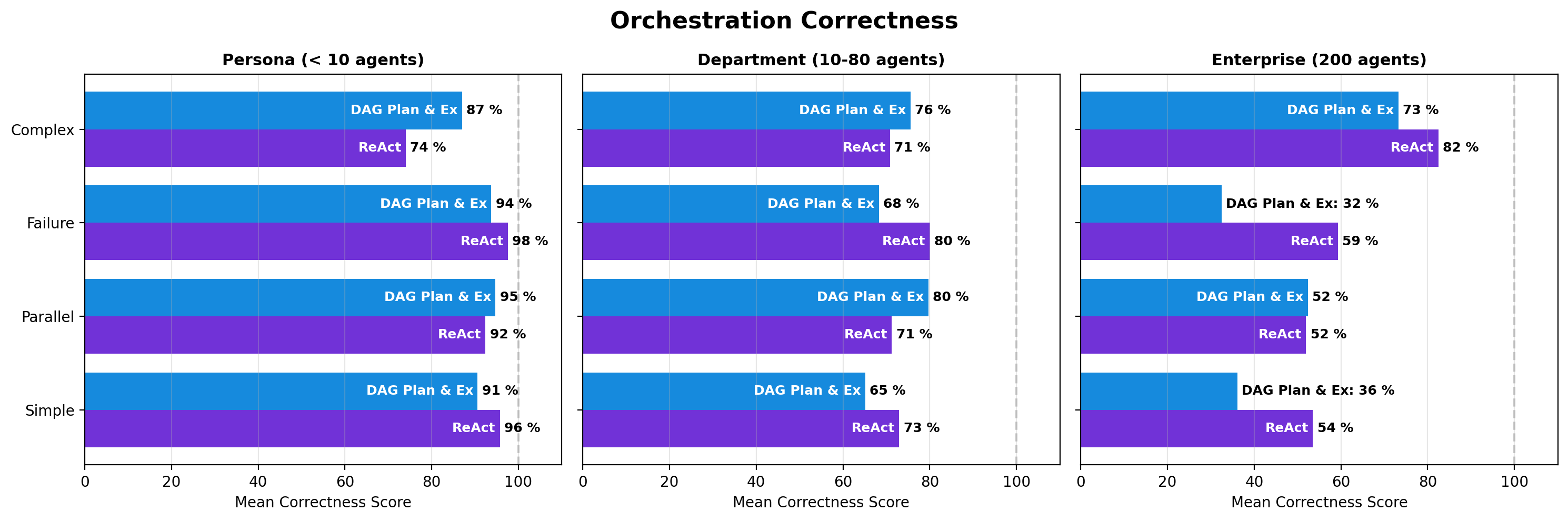}
\caption{Mean correctness scores by scenario type and organizational scale.}\label{fig:correctness-scores}
\end{figure}
The dataset includes 95 Persona-level, 63 Department-level, and 50 Enterprise-level scenarios. Each scenario specifies: a natural language user prompt, the expected agents and execution pattern (sequential, parallel, or mixed), per-agent guidelines including expected identifiers, keywords, and ground-truth response data, and expected answer fragments used for evaluation. This structure enables automated scoring of agent discovery precision/recall, execution pattern correctness, entity extraction accuracy, and response completeness. Dataset construction and release constraints are described in Appendix~\ref{app:dataset-construction} and Appendix~\ref{app:reproducibility}; the scenario schema and examples appear in Appendix~\ref{app:scenario-schema}, and the sanitized research agent catalog appears in Appendix~\ref{app:agent-catalog}.

\section{Results}
\subsection{Orchestration}
\begin{figure}[!tbp]
\centering
\includegraphics[alt={Agent-call precision and recall for DAG Plan and Execute and ReAct, split by scenario type and organizational cohort.},width=0.74\linewidth,trim=0 8bp 0 6bp,clip]{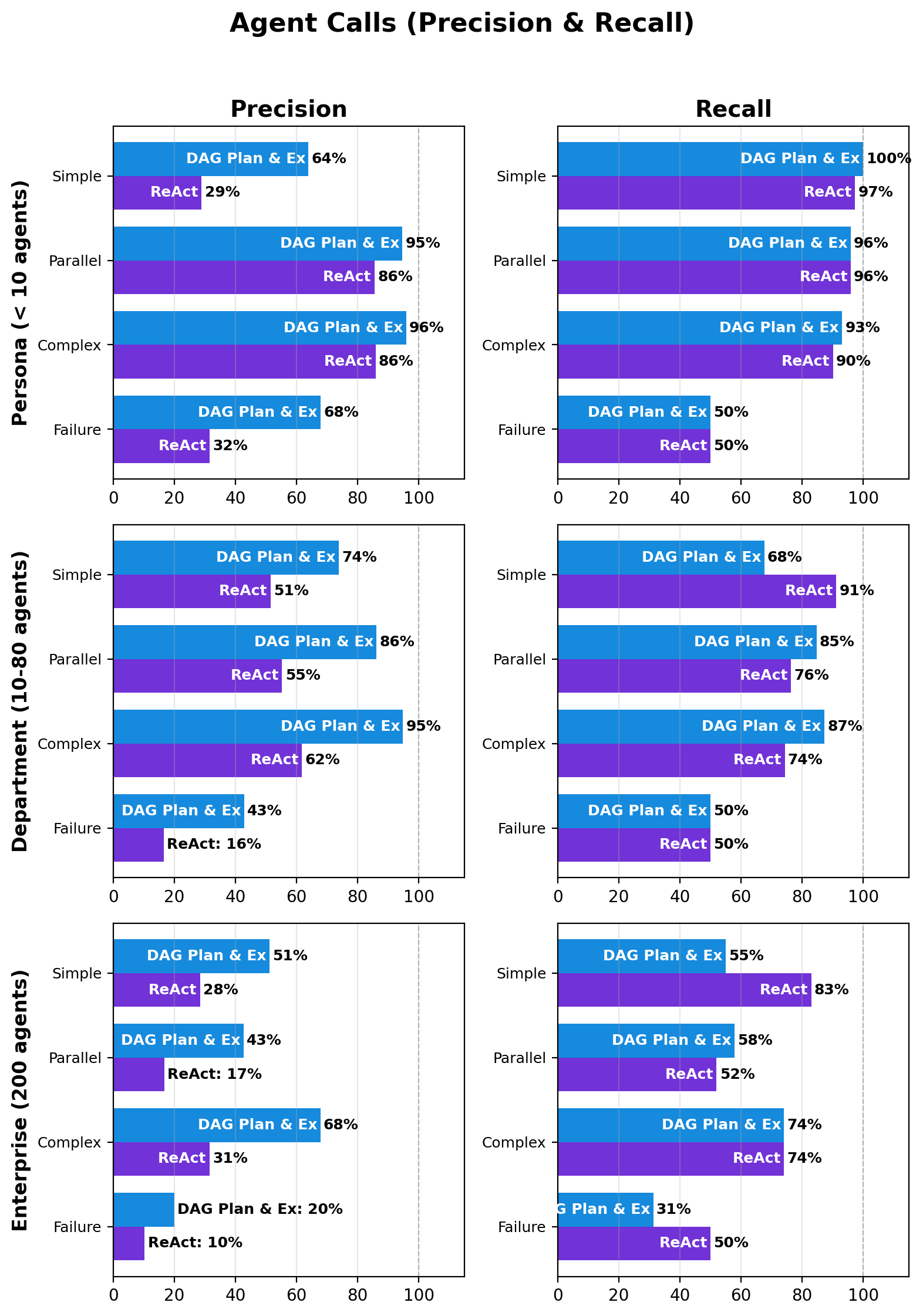}
\caption{Agent call precision and recall by scenario type and organizational scale.}\label{fig:precision-recall}
\end{figure}
\begin{figure}[!t]
\centering
\includegraphics[alt={Expected Answer Completeness for related-event scenarios with and without the Task Manager across organizational cohorts.},width=\linewidth]{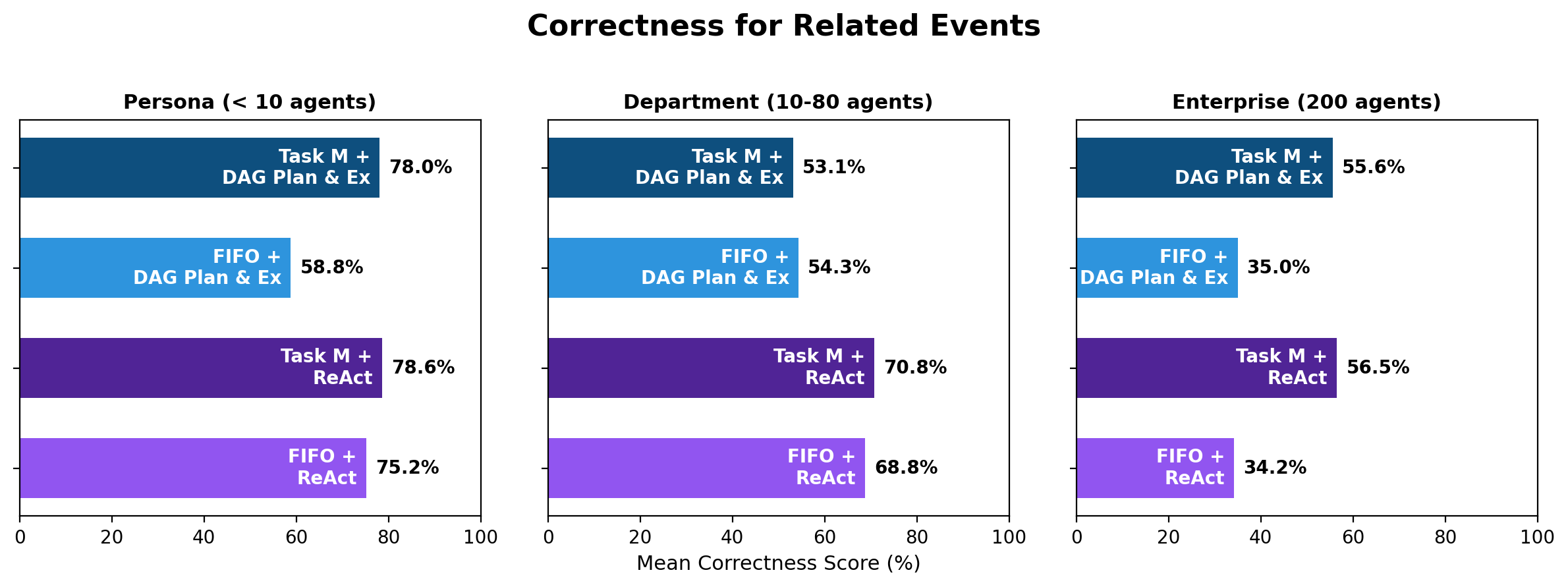}
\caption{Correctness scores for related event scenarios by organizational scale.}\label{fig:steering-correctness}
\end{figure}

Figure~\ref{fig:correctness-scores} shows Expected Answer Completeness (labeled \emph{Correctness} in the figure) across scenario types and organizational scales. Both tested systems score 74--98\% on the Persona cohort, 65--80\% on the Department cohort, and 32--82\% on the Enterprise cohort. \textbf{Simple scenarios show the largest cohort difference}: scores fall from 91--96\% at Persona to 36--54\% at Enterprise, while Complex scores are 74--87\% and 73--82\%, respectively. A larger and more confusable discovery space is one plausible explanation, but the cohorts contain different scenarios, so this comparison does not establish a causal effect of registry size.

\textbf{Enterprise comparison.} At Enterprise scale, the tested ReAct system has higher Expected Answer Completeness on Complex (82\% vs 73\%), Failure (59\% vs 32\%), and Simple (54\% vs 36\%) scenarios; Parallel is tied at 52\%. The tested DAG system's Failure score decreases from 93.7\% at Persona to 68.3\% at Department and 32.4\% at Enterprise. These are whole-system comparisons: the implementations differ in discovery timing, planning, execution, and recovery policy, so the observed differences cannot be assigned to the DAG or ReAct abstraction alone.

\textbf{Precision and recall analysis (Figure~\ref{fig:precision-recall}).} The tested DAG system has higher agent-call precision across the reported cells, while the tested ReAct system has higher recall in several Enterprise conditions. These are descriptive patterns rather than an isolated architectural mechanism because the two systems also differ in discovery timing, planning, execution, and recovery. Confidence intervals in Appendix~\ref{app:extended-results} show substantial across-scenario variation in the Enterprise cohort; Expected Answer Completeness intervals span roughly $\pm$11--30 percentage points, depending on system and scenario type.

\subsection{Event Management}

The Task Manager adds two measurable event-stream capabilities to the request-response orchestrators: merging related events into unified tasks, and priority-based preemption. We evaluate each independently and interpret the results as evidence about continuous-workload behavior in the controlled harness.

{\looseness=-1
\textbf{Related events (Figure~\ref{fig:steering-correctness})} test whether the system can identify semantically connected events--such as a sensor alert followed by a maintenance request for the same equipment--and merge them into a single task that leverages cross-event context. Without merging, each event is processed independently, missing shared entity identifiers and situational awareness. Task Manager improves Expected Answer Completeness at Enterprise scale by over 20 percentage points for both architectures: ReAct rises from 34\% to 57\%, DAG from 35\% to 56\%. At Persona scale, DAG rises from 59\% to 78\%; fewer concurrent events may contribute, but this evaluation does not establish causality.
\par}
\clearpage

\textbf{Priority handling (Table~\ref{tab:priority-queue-time})} measures queue time for high-priority events arriving while lower-priority work is in progress. Events arrive without priority labels--the Task Manager infers priority from event content and context (ground-truth priority was maintained only for evaluation). Without preemption, FIFO processing forces critical events to wait behind long-running tasks. In this harness, DAG shows especially high wait times at Department and Enterprise levels. Those runs include long-running structured plans, but the evaluation does not isolate their contribution to wait time. Task Manager preemption keeps queue time more stable across scale, saving 14--75\% of wait time across both architectures. Because mock agents respond faster than production agents, these queue-time numbers should be read as controlled evidence about relative scheduling behavior rather than as production latency estimates.

\begin{table}[!tbp]
\centering
\scriptsize
\setlength{\tabcolsep}{3pt}
\renewcommand{\arraystretch}{0.92}
\begin{tabular}{@{}lccc@{}}
\toprule
System & Persona & Department & Enterprise \\
\midrule
FIFO + ReAct & 21s & 79s & 107s \\
Task M + ReAct & 15s & 68s & 56s \\
Saved (ReAct) & 29\% & 14\% & 48\% \\
FIFO + DAG & 108s & 323s & 305s \\
Task M + DAG & 42s & 81s & 79s \\
Saved (DAG) & 61\% & 75\% & 74\% \\
\bottomrule
\end{tabular}
\caption{Mean queue time for high-priority events and percentage reduction with Task Manager.}\label{tab:priority-queue-time}
\end{table}

Additional related-event distributions, token usage, and cost analyses are reported in Appendix~\ref{app:extended-results}.

\section{Enterprise Deployment Implications}
In enterprise deployments, the choice between DAG Plan \& Execute and ReAct should be driven less by task complexity than by registry scale and recovery cost. DAG Plan \& Execute is attractive when the relevant agent set is small or moderately sized, dependencies are known, and parallelization can be exploited. Its upfront planning improves precision and makes execution structure explicit. However, the experiments show that at enterprise scale, repeated discovery inside plan-interrupt-replan loops can compound errors and increase latency.

ReAct is the stronger default when discovery is noisy, failure recovery must be incremental, or the cost of replanning is high. Its stepwise interaction gives it fewer opportunities to overcommit to an incorrect global plan. This makes it more robust at 200-agent scale, especially on failure scenarios where a structured replanning architecture repeatedly queries the same noisy discovery layer.

The Task Manager is most valuable when the workload is continuous rather than request-response. Priority inference, related-event merging, and preemption convert an orchestrator from a single-task assistant into a scheduler for event streams. In practice, enterprise systems should monitor backlog age, preemption frequency, merge accuracy, repeated agent calls, and discovery recall. These signals reveal whether the bottleneck is agent selection, task scheduling, agent response quality, or synthesis.

This evaluation uses a single orchestration worker, but enterprise deployments will likely need multiple workers. We do not study distributed scheduling here. However, the design choice that each backlog task is independent is important for that setting: independent tasks can be partitioned, leased, or migrated across workers without requiring every worker to share the full conversational state of every other task. The difficult open problem is then coordination around priority ordering, preemption, and fairness across partitions rather than the representation of the task itself.

Multi-user deployments add a second scaling problem. Different users subscribe to different event streams based on their responsibilities and privileges, so user-scoped events should be processed from that user's permission context. Company-wide or department-wide broadcasts are different: naively running every broadcast through every user's orchestrator can waste tokens and compute while repeating the same work. A practical pattern is to process broad broadcasts only when they merge with a user-scoped event, giving the orchestrator the user's perspective and objective on the shared signal. Separately, a general organizational orchestrator can process company-level broadcasts once for aggregate awareness. This avoids rerunning identical tasks unless a new user perspective, privilege context, or objective justifies it. The right policy will depend on organizational structure, job roles, event frequency, and the nature of the work, but event scoping is a necessary design consideration before autonomous event streams are allowed to run continuously.

Finally, the orchestrator should not be the component that decides whether a consequential action is safe. Its role is to schedule work, route context, preserve state, and surface control signals. Safety decisions should remain with the workflow-specific specialist agents that understand the relevant policy, data boundary, and approval rule. For example, an HR compensation agent should enforce pay-equity and manager-approval constraints, a finance close agent should enforce posting thresholds and segregation-of-duties checks, and an IT operations agent should require operator approval before restarting or disabling critical services.

This separation is especially important for human-in-the-loop (HITL) workflows. A HITL request should not be treated as an ordinary model failure or silently resolved by the orchestrator. The orchestrator should pause the affected task, propagate the approval request to the right human or workflow owner, preserve the task state, and resume only after an authorized response is received. In this deployment pattern, guardrails are both domain-specific and workflow-specific: specialist agents encode the business rules for their domain, while the orchestrator enforces scheduling, context flow, auditability, fallback behavior, and cross-agent coordination.

\section{Conclusion}

We evaluated DAG Plan \& Execute and ReAct across 208 scenarios in a sanitized SAP-derived research snapshot spanning Persona, Department, and Enterprise registry cohorts. Both tested systems achieve lower Expected Answer Completeness on the Enterprise cohort than on the Persona cohort, with Simple scenarios showing the largest difference. A larger and more confusable discovery space is a plausible explanation, but the unmatched cohorts do not isolate registry size as a causal factor. The tested ReAct system has higher Enterprise scores on three scenario types and ties on Parallel, whereas the tested DAG system generally has higher agent-call precision. Because the implementations also differ in discovery timing, planning, execution, and recovery, these results characterize the tested systems rather than DAG and ReAct in general. The Task Manager evaluation shows that a lightweight event-stream layer can improve related-event handling and high-priority responsiveness in a controlled harness, reducing queue time by 14--75\% for high-priority tasks and improving related-event Expected Answer Completeness by more than 20 percentage points at Enterprise scale. These findings are diagnostic evidence from a controlled benchmark rather than a claim of complete enterprise deployment.

\section*{Limitations}

\begin{itemize}[leftmargin=*,itemsep=0.1em,topsep=0.2em,parsep=0pt]
\item \textbf{Controlled Simulation, Not Production Validation.} The benchmark is derived from enterprise workflows and active agent-use-case material, but it does not run against live production agents, production systems, or real users. The results therefore measure orchestration behavior under controlled ground truth rather than deployment readiness, user trust, operational reliability, or business impact.

\item \textbf{Mock Agent Latency.} Mock agents execute a single LLM call, producing responses significantly faster than production agents that perform multi-step reasoning with tool invocations. This compresses queue dynamics and underrepresents the window during which higher-priority events may arrive and trigger preemption, making queue wait times difficult to benchmark directly.

\item \textbf{Limited Artifact Release.} Exact numerical reproduction is constrained by proprietary source material, including original enterprise user stories, workflow specifications, internal agent catalog details, and full per-agent guideline files. We provide the scenario schema, sanitized research catalog, prompts, metrics, judge prompt, aggregate results, and release-scope details to support structural replication rather than exact benchmark replay.

\item \textbf{Fixed Orchestration Parameters.} Tunable parameters such as the checkpoint quantum were held constant to isolate the effect of orchestration topology from configuration tuning. Optimal values are likely workload-dependent, and sensitivity analysis across these dimensions remains as future work.

\item \textbf{No Agentic Memory.} The evaluation excludes persistent memory to establish a controlled baseline independent of learned optimizations. Results therefore represent a lower bound; memory-augmented deployments would likely improve on repeated event patterns.

\item \textbf{Single Orchestration Worker.} All experiments assume one orchestration worker with parallel agent dispatch within each task. Multi-worker deployments introduce backlog partitioning and distributed priority ordering challenges not addressed here.
\end{itemize}

\section*{Ethical Considerations}

\begin{itemize}[leftmargin=*,itemsep=0.1em,topsep=0.2em,parsep=0pt]
\item \textbf{Data Collection \& Consent.} No human annotators or crowd-workers were involved. Evaluation scenarios were based on real enterprise agent use cases, sanitized SAP-derived workflow specifications, and representative user stories. Mock agents simulate enterprise behavior without accessing production systems.

\item \textbf{Biases \& Guardrails.} This work uses existing commercial LLMs; we did not train models ourselves. Priority inference and agent discovery may reflect inherent biases. The orchestrator synthesizes responses from multiple agents, introducing synthesis bias risk--particularly sensitive for HR workflows. Guardrails responsibility is delegated to downstream specialist agents who implement process-specific controls with domain expertise, unlike single-agent systems where guardrails must be one-size-fits-all. Our LLM-as-judge evaluation against known expected answers penalizes synthesis that omits or distorts information.

\item \textbf{Misuse \& Dual-Use.} Multi-agent orchestration at scale could enable malicious coordination. Human-in-the-loop decisions are delegated to specialist agents; the orchestrator propagates HITL requests upward rather than making consequential decisions autonomously. This enables workflow-specific safety controls encoded by process experts. All evaluation used mock agents; no real external systems were affected.

\item \textbf{Environmental Impact.} No model training was conducted; this work uses inference-only evaluation against existing commercial LLM APIs. Computational costs including token usage are reported in Appendix~\ref{app:extended-results}.

\item \textbf{Responsible NLP Checklist Coverage.} A consolidated mapping to the ARR Responsible NLP checklist is provided in Appendix~\ref{app:responsible-nlp-checklist}.
\end{itemize}

\noindent\textbf{AI Use Statement:} This paper was prepared with AI writing assistance tools in accordance with the ACL Policy on AI Writing Assistance. Claude Opus 4.5 and 4.6, GPT 5.5, and GPT 5.6 Sol were used for prose refinement, improving clarity and readability of author-written content. Gemini 3.1 Pro was used as a discovery tool to identify potentially relevant literature, similar to a search engine. All cited works were subsequently retrieved, read, and verified by the authors to ensure accuracy and relevance. The research contributions, system architecture, technical arguments, and all substantive intellectual content are the original work of the authors. Figure~\ref{fig:vision} is AI generated using Nano Banana 2.

\FloatBarrier
\section*{Acknowledgments}
We gratefully acknowledge the colleagues across SAP who supported this work. We thank Wei Ming Neo, Abhijeet Bhanjadeo, Steven Fu, Madhur Mayank Sharma, Michael Still, and Jonathan von Rueden from the SAP Business AI \& Platform team. We also thank SAP Chief Technology Officer Philipp Herzig and Manik Narayan Saha from SAP Labs East Asia, Singapore, for their support and encouragement.
\bibliographystyle{plainnat}
\bibliography{references}
\clearpage
\appendix
\section*{Appendix}
\section{DAG Plan \& Execute: Architectural Details and Design Tradeoffs}\label{app:dag-architecture}

The system architecture draws from operating system kernel design. The Task Manager functions as the process scheduler (priority queue), the Planner as the process loader (execution graph construction), and the Executor as the CPU dispatcher (work execution). A central design decision is to orchestrate pre-existing specialist agents via the A2A protocol rather than dynamically spawning subagents -- trading generalization flexibility for domain-specific accuracy and reliability.

\textbf{Two-Tier Agent Discovery.} Discovery operates in two phases, inspired by Anthropic's Skills system \citep{anthropic2025skills}. In Phase 1, lightweight agent summaries are retrieved for all registered agents, enabling the LLM to select relevant agents without loading full context. In Phase 2, complete agent cards with full skill definitions are loaded only for the selected agents, providing the detail required for DAG construction. The same full-detail cards are presented during both initial planning and replanning for better focus during planning.

\textbf{Executor Statelessness.} The Executor has no access to conversation or orchestration history -- it sees only the current OrchestrationState (DAG, completed nodes, accumulated context, quantum counter). To compensate, the Planner attaches \texttt{processing\_instructions} to DAG edges, carrying institutional memory into the execution graph itself.

\textbf{Quantum Budget.} The quantum limits agent calls before forcing a checkpoint. Two modes: percentage-based (\texttt{quantum\_n < 1}, fraction of DAG size) or absolute (\texttt{quantum\_n $\geq$ 1}, fixed call count). A minimum guarantee prevents degenerate cases on large DAGs.

\textbf{Replanning.} Triggered by CHECKPOINT (quantum exhausted), INTERRUPT (node failure), or RESUME (returning from preemption). A two-level LLM decision determines: (1) whether replanning is needed, and (2) whether different agents are required. When current agents suffice, a DAG edit preserves completed nodes via a replanning boundary marker. When new agents are needed, full discovery reruns with prior context included. Merged events arriving mid-execution are queued and presented at the next checkpoint, enabling dynamic steering.

\textbf{Fast Model Usage.} All components use the primary model except the Executor, which uses the fast model for dispatch-level decisions (interrupt assessment, output evaluation) where the reasoning complexity is low and latency savings compound across many sequential agent calls.

\textbf{Preemption During Agent Waits (Open Problem).} Preemption currently occurs only at quantum boundaries. If urgent work arrives while awaiting an agent response, the system cannot switch until the call completes. Potential solutions include extending A2A polling to simultaneously check the backlog, or factoring accumulated wait time into priority scoring.

\textbf{Runtime-Configurable Parameters.} Merge rules, scoring weights, preemption thresholds, and quantum parameters are designed for runtime adjustment, enabling per-deployment tuning without code changes.

\FloatBarrier
\section{Task Manager: Event Processing and Priority Scoring}\label{app:task-manager-scoring}

The Task Manager processes incoming events through an LLM-based decision function that determines whether each event should spawn a new independent task or merge into an existing one. The following system prompt governs this decision:

\begin{lstlisting}
You are a task manager for an orchestration system.
Your job is to decide how to handle incoming events.

Your options are:
1. "create" -- Create a new independent task for this event.
2. "merge"  -- Merge this event into an existing related task.

CRITICAL RULES:
- All tasks in the backlog must be INDEPENDENT of each other
- Only merge if the new event is clearly related to an existing task's goal
- When in doubt, create a new task
- Assign appropriate priority based on event severity and urgency

PRIORITY GUIDELINES:
- LOW: Informational events, routine maintenance
- MEDIUM: Standard operational events (default)
- HIGH: Important events requiring timely attention

Respond with a JSON object containing your decision.
Make sure to include all important information in the new task description!
\end{lstlisting}

Each event is presented alongside the current backlog state, enabling the model to assess relatedness:

\begin{lstlisting}
EVENT DETAILS:
Type: {event.type}
Source: {event.source}
Data: {event.data}

EXISTING TASKS IN BACKLOG:
{existing_tasks_desc}

Analyze this event and decide:
1. Should this create a new task or merge into an existing one?
2. What priority should it have?
3. What should the task request/goal be?

Remember: Tasks must be independent. Only merge if clearly related.
\end{lstlisting}

The model returns a structured JSON response conforming to the following schema:

\begin{lstlisting}
{
  "type": "object",
  "required": ["decision", "task_request", "priority", "reason"],
  "properties": {
    "decision": {
      "type": "string",
      "enum": ["create", "merge", "ignore"],
      "description": "Whether to create a new task, merge into an existing one, or ignore this event"
    },
    "merge_into_task_id": {
      "type": ["string", "null"],
      "description": "If decision is 'merge', the ID of the task to merge into"
    },
    "task_request": {
      "type": "string",
      "description": "The task goal/request description (for new task or updated merged task). No information from the original request should be lost."
    },
    "priority": {
      "type": "string",
      "enum": ["low", "medium", "high"],
      "description": "The priority level for the task"
    },
    "reason": {
      "type": "string",
      "description": "Explanation for the decision"
    }
  }
}
\end{lstlisting}

Decisions are validated against configuration constraints. If validation fails, the system re-prompts with error feedback or falls back to safe defaults, ensuring no event is dropped.

Tasks in the backlog are ranked by a composite priority score:

\[
\begin{aligned}
\text{score} ={}& \text{base\_priority} \times
(1.0 + \text{time\_bonus}) \\
&+ \text{progress\_factor}
\end{aligned}
\]

where $\text{base\_priority} \in \{1.0, 2.0, 3.0\}$ corresponds to LOW, MEDIUM, and HIGH respectively. The time bonus is defined as $\max(0,\; (\text{task\_age} / \text{mean\_age}) - 1.0) \times w_t$ with default weight $w_t = 0.1$, activating only when a task has waited longer than the backlog average. This prevents starvation: a LOW-priority task waiting twice the mean age receives a 10\% boost to its base score, ensuring it eventually rises in the queue. The progress factor is $(\text{completed\_nodes} / \text{total\_nodes}) \times w_p$ with default $w_p = 0.5$, incentivizing completion of in-progress tasks over equally-ranked fresh tasks to reduce context-switching overhead.

\FloatBarrier
\section{Mock Agent Design: Prompt and Classification Logic}\label{app:mock-agent-design}

The mock agent server simulates 200 enterprise agents using a single shared LLM prompt that combines agent identity, story context, and classification rules. Each mock agent receives its identity and domain-specific guidelines at call time, enabling a unified executor to serve all agents without per-agent logic. The following system prompt governs response generation:

\begin{lstlisting}
You are a mock enterprise agent in a test harness.

## Identity
- Agent: {agent_name}
- Description: {agent_description}
- Capabilities: {agent_skills}

## Story Context
Active story: "{story_title}"

## Guidelines
{guidelines_section}

## Response Schema
{response_schema}

## Rules

1. Identify which of YOUR expected_prompts & keywords that the user's prompt relates to.
    - Expected Prompts => Identifier to help identify the entity related to the request
    - Keywords => Information requested about said entity
2. Classify the user's request into the following:
    - MATCH: Can identify the entity (sufficient expected prompts) AND have all requested information (sufficient keywords)
    - CLOSE_MATCH: Can identify the entity (sufficient expected prompts) BUT lack some requested information (insufficient keywords)
    - MISMATCH: Cannot identify the entity (insufficient expected prompts) OR no guidelines apply OR a refusal condition is triggered
3. In case of a MATCH: Return expected_response_info lightly reformatted to match the question where appropriate, but all information must be included.
4. In case of a CLOSE_MATCH: Return the relevant subset of expected_response_info. Do not volunteer extra information that is not asked for.
5. In case of a MISMATCH: Use the appropriate response:
   - Ambiguous entity: "Many records match the <description>. Could you narrow it down?"
   - No applicable guidelines: "This request is outside my area of expertise. I handle <your capabilities>."
   - Refusal triggered: Politely decline
6. NEVER invent data. Only use what's in expected_response_info.
7. Return expected_response_info as-is or the relevant fraction of it. Never add extra information beyond what is provided.

Respond with valid JSON matching the provided schema.
\end{lstlisting}

The classification logic implements entity identification through a two-axis matching system. The first axis (\texttt{expected\_prompts}) identifies which entity the request concerns, while the second axis (\texttt{keywords}) determines whether the agent possesses the requested information about that entity. This produces three outcomes -- MATCH when both axes are satisfied, CLOSE\_MATCH when only entity identification succeeds, and MISMATCH when entity identification fails or a refusal condition triggers.

The structured response conforms to the following schema:

\begin{lstlisting}
{
  "type": "object",
  "required": ["expected_prompts_matched", "keywords_matched", "classification", "reasoning", "response"],
  "properties": {
    "expected_prompts_matched": {
      "type": "array",
      "items": {"type": "string"},
      "description": "Which items from the expected prompts list the user's prompt mentions"
    },
    "keywords_matched": {
      "type": "array",
      "items": {"type": "string"},
      "description": "Which keywords from the keyword list the user's prompt asks about or relates to"
    },
    "classification": {
      "type": "string",
      "enum": ["MATCH", "CLOSE_MATCH", "MISMATCH"],
      "description": "MATCH: request can be addressed using available data. CLOSE_MATCH: partially relevant but cannot fully help. MISMATCH: no keywords match OR no guidelines exist OR refusal trigger hit."
    },
    "reasoning": {
      "type": "string",
      "description": "Brief (1-2 sentences): which expected_prompts and keywords matched and why"
    },
    "response": {
      "type": "string",
      "description": "Agent response using data from expected_response_info. Include all relevant context. Do not invent data. If MISMATCH, politely decline stating capabilities."
    }
  }
}
\end{lstlisting}

This design ensures deterministic, reproducible agent behavior: given identical guidelines and user prompts, the classification outcome is consistent across runs, enabling reliable evaluation of orchestrator routing decisions.

\FloatBarrier
\section{Dataset Construction Methodology}\label{app:dataset-construction}

\begin{figure}[!tp]
\centering
\includegraphics[alt={Human-in-the-loop dataset construction pipeline from sanitized source material through LLM generation, human review, revision loops, and final scenarios.},width=\linewidth]{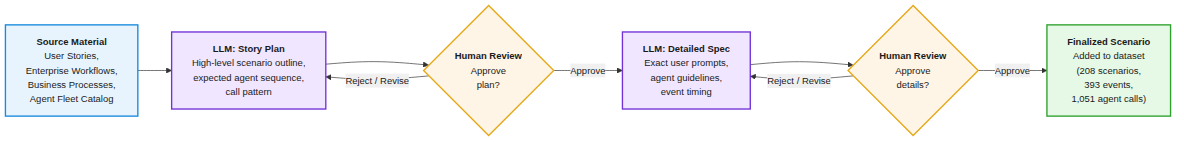}
\caption{Dataset construction pipeline. Blue nodes indicate source material, purple nodes indicate LLM generation stages, gold diamonds indicate human review gates (with revision loops), and the green node indicates a finalized scenario.}\label{fig:dataset-construction}
\end{figure}

The 208 evaluation scenarios were constructed through a human-in-the-loop generation pipeline (Figure~\ref{fig:dataset-construction}). Source material -- comprising sanitized user stories derived from enterprise agent work at SAP SE, documented workflows, and business process specifications -- was provided to an LLM which generated high-level story plans identifying the scenario type, expected agent sequence, and call pattern. Each plan was reviewed by a human with domain knowledge; rejected plans were revised and resubmitted until approved. Approved plans then entered a second generation stage where the LLM produced exact specifications: natural language user prompts, per-agent guidelines (expected\_prompts, keywords, expected\_response\_info), event timing and dependencies. These details underwent a second human review loop to verify correctness of entity boundaries, response fidelity, and scenario completeness before finalization into the evaluation dataset.

\section{Reproducibility and Release Scope}\label{app:reproducibility}

Because this evaluation is based on sanitized enterprise agent use cases assembled at SAP SE, we cannot release the raw source material used to construct the benchmark. This includes the original user stories, workflow specifications, internal agent catalog material, full per-agent guideline files, and implementation code. Releasing these artifacts could disclose proprietary product information. We therefore separate exact artifact release from methodological reproducibility.

To support reproducibility at the methodological level, the paper reports the benchmark structure and evaluation protocol in detail. We provide the scenario taxonomy, scale definitions, construction pipeline, anonymized agent catalog, scenario schema, representative scenario examples, orchestration prompts, ReAct prompt, LLM-judge prompt, metrics, and aggregate results with confidence intervals. The scenario schema specifies the fields needed to recreate an equivalent benchmark: natural-language event prompts, expected agent sets, expected call patterns, timeout and delay fields, priority labels where applicable, and per-agent expected response fragments. These materials are sufficient to reproduce the evaluation design, but not to replay the proprietary benchmark exactly.

An external replication can therefore reproduce the evaluation design without the proprietary source artifacts by constructing a sanitized research agent catalog at the same three scales, authoring scenarios under the five scenario types described in Appendix~\ref{app:scenario-schema}, implementing deterministic mock agents whose responses are gated by expected prompt keywords, and scoring orchestration outputs using the metrics and judge prompt reported here. Such a replication would test the same orchestration properties: discovery under scale, sequential dependency handling, parallel dispatch, retry after incomplete responses, mid-execution steering, related-event merging, and priority-based preemption.

Exact numerical reproduction is not expected without the original enterprise workflows and agent catalog. The intended reproducibility target is structural and methodological: another researcher should be able to instantiate a comparable benchmark, run the same classes of orchestration systems, and evaluate them with the same metrics and prompting protocol.

\section{Responsible NLP Checklist Coverage}\label{app:responsible-nlp-checklist}

The paper addresses the ARR Responsible NLP checklist as follows. Limitations are discussed in the Limitations section. Potential risks are discussed in Ethical Considerations, including bias, guardrails, misuse, dual use, and environmental impact. Scientific artifacts are documented through the dataset construction methodology, scenario schema, agent catalog, prompts, and release-scope statement in Appendix~\ref{app:dataset-construction}, Appendix~\ref{app:scenario-schema}, Appendix~\ref{app:agent-catalog}, Appendix~\ref{app:dag-prompts}, Appendix~\ref{app:react-prompt}, and Appendix~\ref{app:judge-prompt}. We do not release raw enterprise source material, internal agent catalog material, full per-agent guideline files, or implementation code because they contain proprietary product information; instead, we provide the structure needed for methodological replication in Appendix~\ref{app:reproducibility}. No human annotators, crowd-workers, or recruited study participants were used. Human review in the scenario-construction pipeline was performed by domain-knowledgeable authors or internal product experts as part of benchmark design, not as a human-subjects study. No model training was conducted; experiments used inference-only commercial LLM APIs and deterministic mock agents. The evaluation setup, metrics, confidence intervals, token usage, and cost estimates are reported in the main paper and Appendix~\ref{app:extended-results}. AI writing-assistance use is disclosed in the AI Use Statement.

\FloatBarrier
\section{Scenario Structure and Type Taxonomy}\label{app:scenario-schema}

The evaluation dataset comprises 208 scenarios organized into five types: complex, parallel, failure, steering, and priority. Each scenario is defined as a structured document containing metadata, one or more events (user prompts that trigger orchestration), and agent guidelines (expected behavior contracts for mock agents). During evaluation, only the event's \texttt{user\_prompt} is passed to the system under test -- the orchestrator must independently discover agents, plan execution order, and synthesize prompts without access to the ground-truth fields. This section first presents the full schema through a representative example, then illustrates each type's distinguishing characteristics.

\subsection{Part 1: Full Scenario Anatomy}

The following is a representative \emph{complex} scenario exercising a six-agent sequential pipeline for a new product introduction workflow.

\textbf{Scenario Metadata:}

\begin{itemize}[leftmargin=*,itemsep=0.1em,topsep=0.2em,parsep=0pt]
\item ID: \texttt{department-supply-chain-complex-001}.
\item Title: New Product Introduction End-to-End.
\item Type: \texttt{complex}.
\item Business domain: Supply Chain.
\item Level: Department (cross-functional).
\item Description: full lifecycle from recipe development through supplier sourcing, qualification, production scheduling, shop floor preparation, and asset readiness verification. Tests orchestrator routing across formulation, procurement, and manufacturing agents in sequence.
\end{itemize}

\textbf{Event:}

\begin{itemize}[leftmargin=*,itemsep=0.1em,topsep=0.2em,parsep=0pt]
\item ID: \texttt{evt-001}.
\item Call pattern: sequential.
\item Timeout: 60 seconds.
\item User prompt: \emph{"We are introducing a new high-performance thermal paste product targeting the semiconductor packaging market. Execute the full NPI process: develop the recipe formulation for a silicone-free thermal interface material, identify and qualify suppliers for boron nitride platelets and aluminium oxide nanoparticles, onboard the selected supplier from South Korea, create the production plan for an initial batch at our facility, prepare the shop floor including workstation setup and operator training, and verify that our mixing equipment can handle the new formulation parameters."}
\end{itemize}

\textbf{Expected Agent Sequence:}

\begin{itemize}[leftmargin=*,itemsep=0.1em,topsep=0.2em,parsep=0pt]
\item \texttt{recipe-formulation-agent}: develops material formulation meeting thermal and viscosity specifications.
\item \texttt{supplier-discovery-agent}: identifies qualified suppliers for raw materials.
\item \texttt{supplier-onboarding-agent}: qualifies and registers the selected new vendor.
\item \texttt{production-planning-operations-agent}: creates batch production schedule and BOM.
\item \texttt{shop-floor-supervisor-agent}: prepares workstations, training, and quality checkpoints.
\item \texttt{asset-health-agent}: verifies equipment readiness for new formulation parameters.
\end{itemize}

\textbf{Agent Guidelines (example -- \texttt{recipe-formulation-agent}):}

\begin{itemize}[leftmargin=*,itemsep=0.1em,topsep=0.2em,parsep=0pt]
\item Expected prompts: \texttt{thermal paste}, \texttt{silicone-free}, \texttt{8 W/mK}, \texttt{viscosity}, \texttt{boron nitride}.
\item Keywords: \texttt{formulation}, \texttt{thermal conductivity}, \texttt{silicone-free}.
\item Expected response: Recipe Formulation -- TP-ULTRA-500; Component 1: Hexagonal Boron Nitride platelets (62\% by weight); Component 2: Aluminium Oxide nanoparticles (8\%); Component 3: Polyol ester base fluid (26\%); predicted thermal conductivity: 8.4 W/mK; processing: 4-stage mixing with 48-hour cure.
\end{itemize}

\emph{(Guidelines for remaining 5 agents follow the same structure, each specifying domain-specific expected prompts, keywords, and ground-truth response content.)}

\textbf{Schema Field Definitions:}

\begin{itemize}[leftmargin=*,itemsep=0.1em,topsep=0.2em,parsep=0pt]
\item \texttt{story\_id}: unique identifier encoding level, department, type, and sequence number.
\item \texttt{title}: human-readable scenario name.
\item \texttt{description}: summary of what the scenario tests in the orchestrator.
\item \texttt{type}: one of \texttt{complex}, \texttt{parallel}, \texttt{failure}, \texttt{steering}, or \texttt{priority}.
\item \texttt{department}: business domain, such as \texttt{supply\_chain} or \texttt{finance}.
\item \texttt{level}: scenario scope, either \texttt{persona}, \texttt{department}, or \texttt{enterprise}.
\item \texttt{events[].id}: event identifier within the scenario.
\item \texttt{events[].user\_prompt}: natural-language request submitted to the orchestrator.
\item \texttt{events[].expected\_agents\_called}: ground-truth agent list the orchestrator should invoke.
\item \texttt{events[].expected\_call\_pattern}: expected topology: \texttt{sequential}, \texttt{parallel}, or \texttt{any\_order}.
\item \texttt{events[].timeout\_seconds}: maximum allowed orchestration time for this event.
\item \texttt{events[].delay\_seconds}: delay before injecting this event, used in multi-event scenarios.
\item \texttt{events[].depends\_on}: causal dependency on a prior event.
\item \texttt{events[].priority}: explicit priority level: \texttt{low}, \texttt{medium}, or \texttt{high}.
\item \texttt{agent\_guidelines\{\}}: per-agent behavior contracts specifying expected prompts, response content, and keywords.
\item \texttt{agent\_guidelines.expected\_prompts}: keywords the mock agent uses to classify incoming requests as MATCH or MISMATCH.
\item \path{agent_guidelines.expected_response_info}: substantive content the mock agent returns when classification is MATCH.
\item \texttt{agent\_guidelines.keywords}: additional terms for fuzzy matching.
\end{itemize}

The \texttt{agent\_guidelines} mechanism enables deterministic evaluation: mock agents do not generate arbitrary responses but instead pattern-match on the orchestrator's synthesized prompt and return pre-specified domain content. This isolates orchestrator behavior (planning, routing, prompt synthesis) from downstream generation quality.

\subsection{Part 2: Type Comparison}

Each scenario type targets a distinct orchestration capability. The examples below are condensed to highlight structural differences.

\textbf{Complex} -- Sequential multi-agent routing where each step depends on the output of the previous.

\begin{itemize}[leftmargin=*,itemsep=0.1em,topsep=0.2em,parsep=0pt]
\item Title: New Product Introduction End-to-End.
\item Events: 1.
\item Agents: 6, sequential.
\item Agent sequence: recipe-formulation to supplier-discovery to supplier-onboarding to production-planning to shop-floor-supervisor to asset-health.
\end{itemize}

\emph{Distinguishing feature:} Single event, multiple agents, strict sequential ordering. The orchestrator must infer data dependencies (e.g., supplier onboarding requires discovery results) and build a DAG that serializes all six calls.

\textbf{Parallel} -- Simultaneous dispatch of independent agent calls.

\begin{itemize}[leftmargin=*,itemsep=0.1em,topsep=0.2em,parsep=0pt]
\item Title: Daily AR Operations Dashboard Refresh.
\item Events: 1.
\item Agents: 4, parallel.
\item User prompt: \emph{"For company code 2000, give me today's AR operations snapshot: top-priority work items by revenue impact, current aging breakdown with predicted cash inflows, yesterday's auto-clearing results, and overdue promise-to-pay commitments."}
\item Agent set: ar-worklist-priority, aging-analysis, smart-clearing, promise-to-pay.
\end{itemize}

\emph{Distinguishing feature:} Single event, multiple agents, all calls are independent. The orchestrator should recognize that none of the four sub-tasks depends on another and execute them concurrently.

\textbf{Failure} -- Detection of incomplete responses and orchestrator-driven retry with increased specificity.

\begin{itemize}[leftmargin=*,itemsep=0.1em,topsep=0.2em,parsep=0pt]
\item Title: Asset Health Assessment Omits Predictive Failure Timeline.
\item Events: 1.
\item Agents: 2, with the same agent called twice sequentially.
\item User prompt: \emph{"We are planning a 2-week continuous production campaign. Give me the full reliability picture -- current condition, degradation trends, and your confidence that it will complete without failure."}
\item Agent calls: asset-health-agent to asset-health-agent.
\end{itemize}

\emph{Distinguishing feature:} The same agent appears twice in the expected call list. The mock agent is configured to return only current-condition metrics on the first call (deliberately omitting predictive analysis). The orchestrator must detect the incomplete response, formulate a more specific retry prompt, and call the same agent again -- the second call triggers the full response including failure probability and maintenance recommendations.

\textbf{Steering} -- Mid-execution redirection when new constraints arrive after work has begun.

\begin{itemize}[leftmargin=*,itemsep=0.1em,topsep=0.2em,parsep=0pt]
\item Title: Revenue Recognition Policy Change Mid-Close.
\item Events: 2.
\item Timing: Event 2 arrives 25 seconds after Event 1 (\texttt{depends\_on: evt-001}).
\item \texttt{evt-001}: start the May 2026 close by posting accruals, generating analytical insights, and preparing consolidation; expected agents are accruals, analytical-insights, and consolidation; pattern is sequential.
\item \texttt{evt-002}: auditors require switching to over-time recognition; reverse accruals, regenerate consolidation, and draft IFRS disclosure; expected agents are accruals, consolidation, and public-disclosure-narrative; pattern is sequential.
\end{itemize}

\emph{Distinguishing feature:} Multiple events with temporal separation (\texttt{delay\_seconds: 25}) and causal dependency (\texttt{depends\_on}). The second event arrives while the first is still being processed, forcing the orchestrator to incorporate new constraints into ongoing work -- reversing already-completed steps and redirecting agent activity.

\textbf{Priority} -- Preemption scheduling when events of varying urgency arrive concurrently.

\begin{itemize}[leftmargin=*,itemsep=0.1em,topsep=0.2em,parsep=0pt]
\item Title: Catastrophic Compressor Failure During Routine Maintenance Planning.
\item Events: 4, staggered over 60 seconds.
\item \texttt{evt-001} at 0s, low priority: plan Q4 preventive maintenance for 9 steam turbines; expected agent is maintenance-request-agent.
\item \texttt{evt-002} at 40s, high priority: main air compressor failed, nitrogen supply is failing in 18 minutes, and immediate response is required; expected agents are alert-processing, maintenance-request, and technician-mobile.
\item \texttt{evt-003} at 55s, low priority: routine health check on 6 cooling tower fans; expected agent is asset-health-agent.
\item \texttt{evt-004} at 60s, medium priority: optimize technician rotation for 6 PMs; no required specialist agent.
\end{itemize}

\emph{Distinguishing feature:} Four events arriving at staggered intervals with explicit priority levels. The Task Manager must preempt in-progress low-priority work when the high-priority emergency arrives at T+40s, service it first, then resume remaining tasks by priority. Event 3 also serves as a minimal single-agent scenario (one event, one agent, routine request) -- the simplest unit of work within the evaluation framework.

\textbf{Summary.} The five types collectively evaluate the core orchestration capabilities: sequential dependency resolution (complex), concurrent dispatch (parallel), response quality detection and autonomous retry (failure), dynamic replanning under new constraints (steering), and priority-based preemption with multi-event scheduling (priority). Every scenario in the dataset conforms to this schema, enabling automated scoring against ground-truth agent lists, call patterns, and timing constraints.

\FloatBarrier
\section{Extended Results Analysis}\label{app:extended-results}

\textbf{Unique-agent precision and recall.} For each scenario, let $E$ be the set of distinct expected agent identities and $A$ the set of distinct invoked agent identities. We compute $\mathrm{Precision}=|A\cap E|/|A|$ and $\mathrm{Recall}=|A\cap E|/|E|$ for non-empty denominators. Repeated occurrences of an agent collapse on both sides; for example, an expected list $[A,A]$ is represented by $E=\{A\}$. This is an agent-identity coverage metric, not a count-sensitive call-sequence metric.

\begin{figure}[!tp]
\centering
\includegraphics[alt={Scenario-level mean agent-call precision and recall with 95 percent confidence intervals by system, scenario type, and cohort.},width=\linewidth]{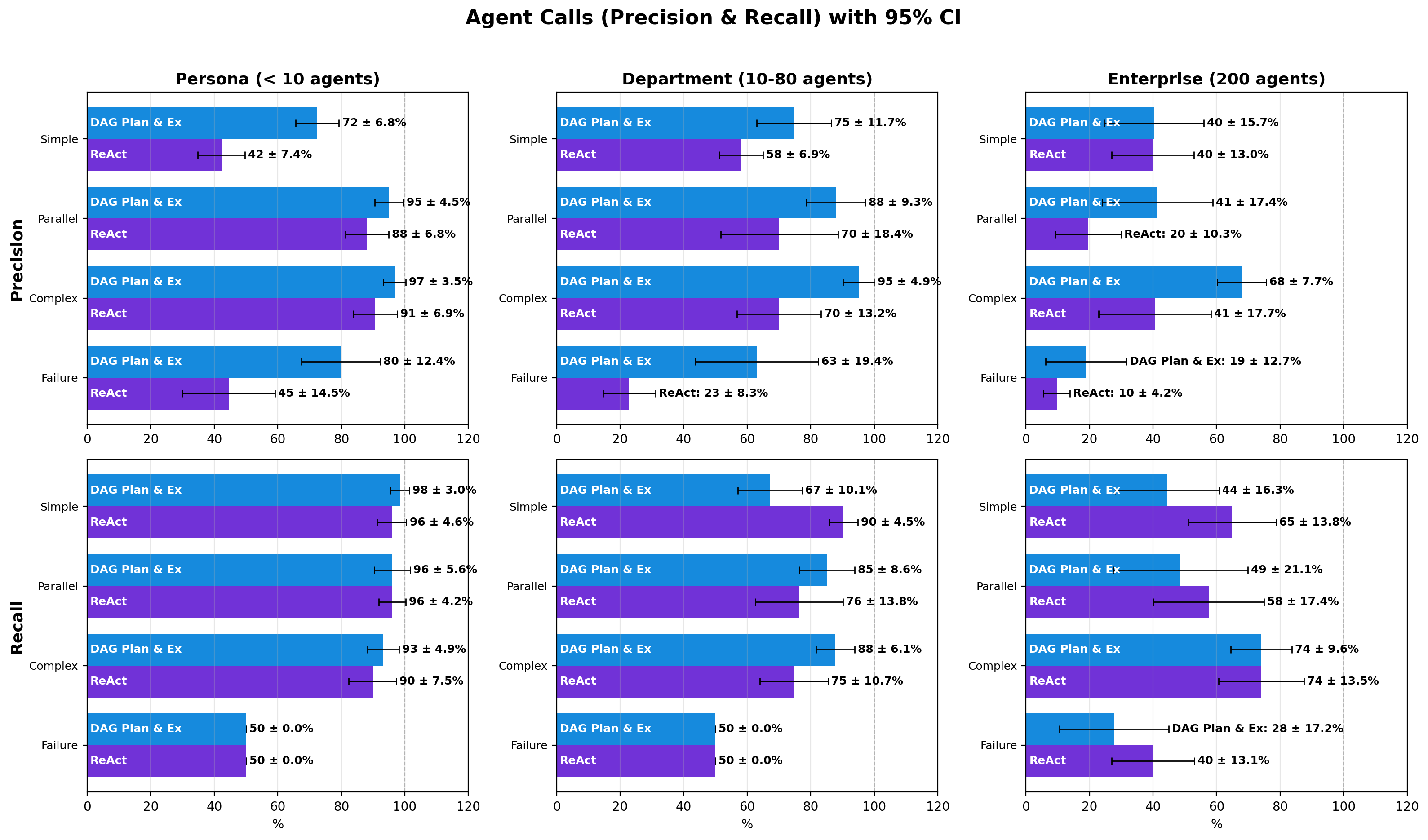}
\caption{Agent call precision and recall by scenario type with 95\% confidence intervals.}\label{fig:precision-recall-ci}
\end{figure}

\textbf{Precision and recall with confidence intervals -- orchestration scenarios (Figure~\ref{fig:precision-recall-ci}).} Figure~\ref{fig:precision-recall-ci} reports scenario-level means with 95\% normal-approximation intervals computed as $1.96s/\sqrt{n}$, where $s$ is the sample standard deviation across scenarios and $n$ is the number of scenarios in the cell. The intervals therefore describe across-scenario variation in this benchmark, not within-task stochastic uncertainty. Enterprise intervals are wider in several cells, including DAG precision for Simple ($\pm$15.7\%) and ReAct recall for Parallel ($\pm$17.4\%). Enterprise Failure recall is 28 $\pm$ 17.2\% for DAG and 40 $\pm$ 13.1\% for ReAct; the interval overlap warrants caution when comparing the tested systems.

\begin{figure}[!tp]
\centering
\includegraphics[alt={Scenario-level mean Expected Answer Completeness with 95 percent confidence intervals by system, scenario type, and cohort.},width=\linewidth]{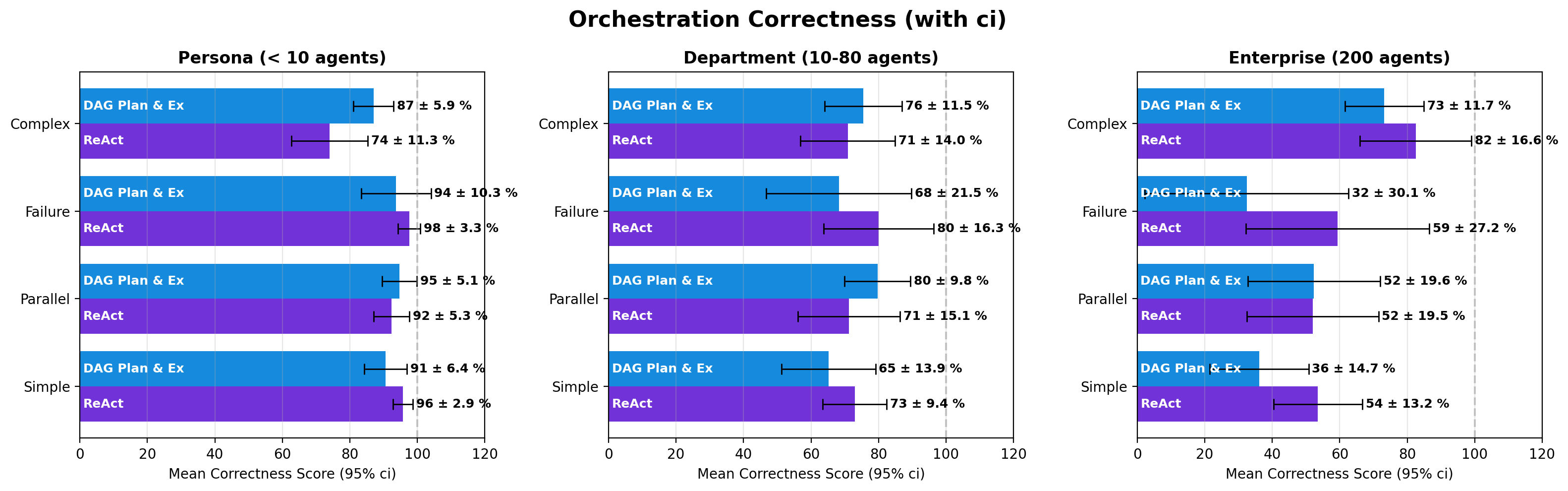}
\caption{Mean correctness score by scenario type for DAG Plan \& Execute vs ReAct across orchestration scales, with 95\% confidence intervals.}\label{fig:correctness-ci}
\end{figure}

\textbf{Expected Answer Completeness with confidence intervals -- orchestration comparison (Figure~\ref{fig:correctness-ci}).} Figure~\ref{fig:correctness-ci} reports scenario-level mean Expected Answer Completeness (labeled \emph{Correctness} in the figure) with 95\% intervals. At Persona, ReAct is higher on Simple (96 $\pm$ 2.9\% vs 91 $\pm$ 6.4\%) and Failure (98 $\pm$ 3.3\% vs 94 $\pm$ 10.3\%), while DAG is higher on Complex (87 $\pm$ 5.9\% vs 74 $\pm$ 11.3\%) and Parallel (95 $\pm$ 5.1\% vs 92 $\pm$ 5.3\%). At Department, DAG is higher in all four displayed cells, including Parallel (80 $\pm$ 9.8\% vs 71 $\pm$ 15.1\%). At Enterprise, ReAct is higher on Complex (82 $\pm$ 16.6\% vs 73 $\pm$ 11.7\%), Failure (59 $\pm$ 27.2\% vs 32 $\pm$ 30.1\%), and Simple (54 $\pm$ 13.2\% vs 36 $\pm$ 14.7\%); Parallel is tied at 52\%. The wider Enterprise intervals ($\pm$11--30\%) indicate greater variation across benchmark scenarios. Because the cohorts are unmatched and the implementations differ in multiple components, these results do not isolate registry size or architecture as a causal factor.

\begin{figure}[!tp]
\centering
\includegraphics[alt={Boxplots of related-event Expected Answer Completeness for Task Manager and FIFO variants across Persona, Department, and Enterprise cohorts.},width=\linewidth]{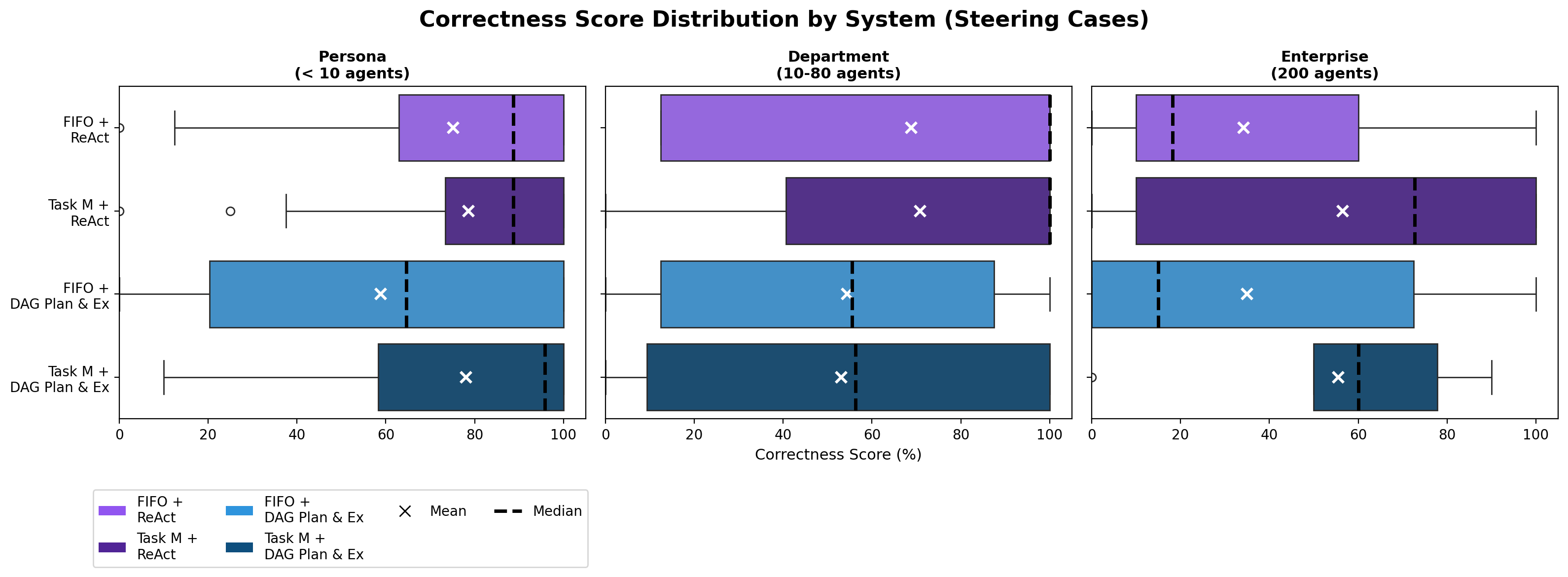}
\caption{Correctness score distribution for related-event scenarios. $\times$ = mean, dashed = median.}\label{fig:steering-boxplot}
\end{figure}

\textbf{Related-tasks Expected Answer Completeness distribution (Figure~\ref{fig:steering-boxplot}).} The boxplot in Figure~\ref{fig:steering-boxplot} shows that the median Expected Answer Completeness exceeds the mean across all system variants, consistent with a small number of low-scoring cases pulling down the averages. At Enterprise scale, Task M+DAG achieves a median near 75\% despite a mean of 56\%, suggesting that when merge decisions succeed, they succeed well, but failures are severe. The wide interquartile ranges at Department level (spanning 20--100\% for all variants) confirm this bimodal behavior: related-event handling either works correctly or fails comprehensively, with few partial successes.

\begin{figure}[!tp]
\centering
\includegraphics[alt={Stacked bars of input and output token usage per event, separated by normal and fast model tiers for each system and cohort.},width=\linewidth]{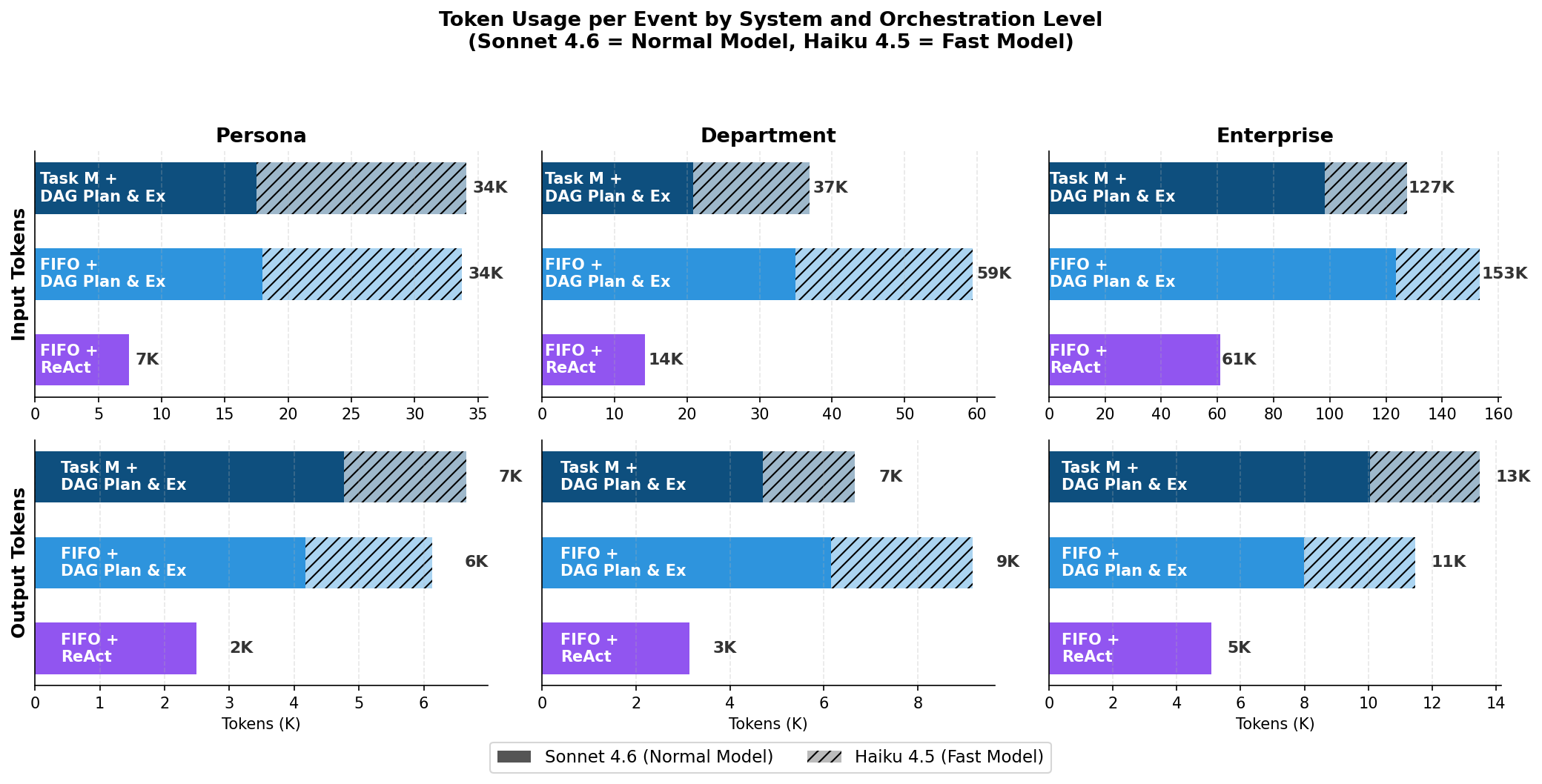}
\caption{Token usage per event by system and orchestration level. Solid bars = Sonnet 4.6 (normal model), hatched bars = Haiku 4.5 (fast model).}\label{fig:token-usage}
\end{figure}

\textbf{Token usage per event (Figure~\ref{fig:token-usage}).} Figure~\ref{fig:token-usage} decomposes token consumption by model tier. DAG Plan \& Execute uses Claude Sonnet 4.6 for normal-model calls and Claude Haiku 4.5 for fast-model calls. From the rounded labels, Persona input-plus-output totals are approximately 41K tokens for Task M+DAG, 40K for FIFO+DAG, and 9K for FIFO+ReAct. Department totals are approximately 44K, 68K, and 17K, respectively; Enterprise totals are approximately 140K, 164K, and 66K. FIFO+ReAct is the lowest-token tested system in every cohort. Because the cohorts contain different scenarios, the increase should be read descriptively rather than as an isolated scaling law.

\begin{figure}[!tp]
\centering
\includegraphics[alt={Mean estimated cost per event with 95 percent confidence intervals for each system and organizational cohort.},width=\linewidth]{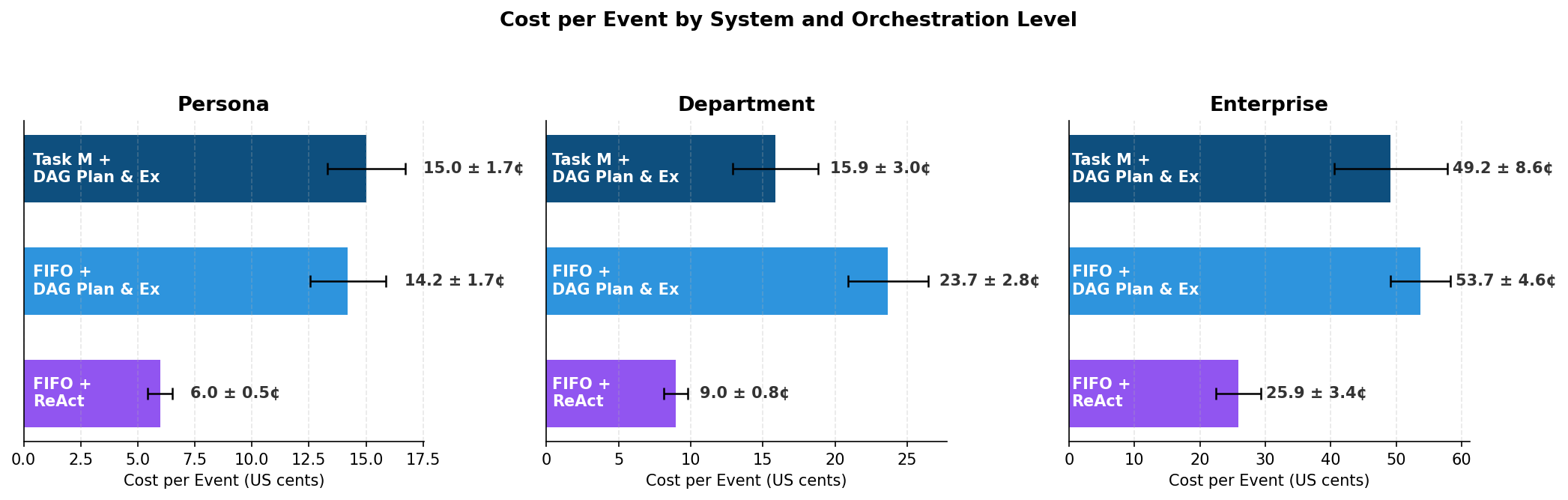}
\caption{Mean cost per event (US cents) with 95\% confidence intervals.}\label{fig:cost-per-event}
\end{figure}

\textbf{Cost per event (Figure~\ref{fig:cost-per-event}).} Figure~\ref{fig:cost-per-event} translates token usage into US-dollar cost estimates using published API pricing (Sonnet 4.6: \$3/M input, \$15/M output; Haiku 4.5: \$1/M input, \$5/M output). FIFO+ReAct is consistently cheapest across all scales (6.0 cents at Persona, 25.9 cents at Enterprise). DAG variants cost 2--2.5$\times$ more at Persona due to planning overhead, but the cost gap narrows at Enterprise scale -- the dual-model split becomes less significant relative to the total observed cost at Enterprise. The 95\% confidence intervals are wider for several Enterprise cells, indicating greater variation across scenarios. Observed mean cost is 3.3--4.3$\times$ higher from Persona to Enterprise, but this descriptive comparison does not estimate a scaling law because the cohorts contain different tasks.

\begin{figure}[!tp]
\centering
\includegraphics[alt={Percentage breakdown of Match, Close Match, and Mismatch classifications for expected-agent responses by orchestrator and cohort.},width=\linewidth]{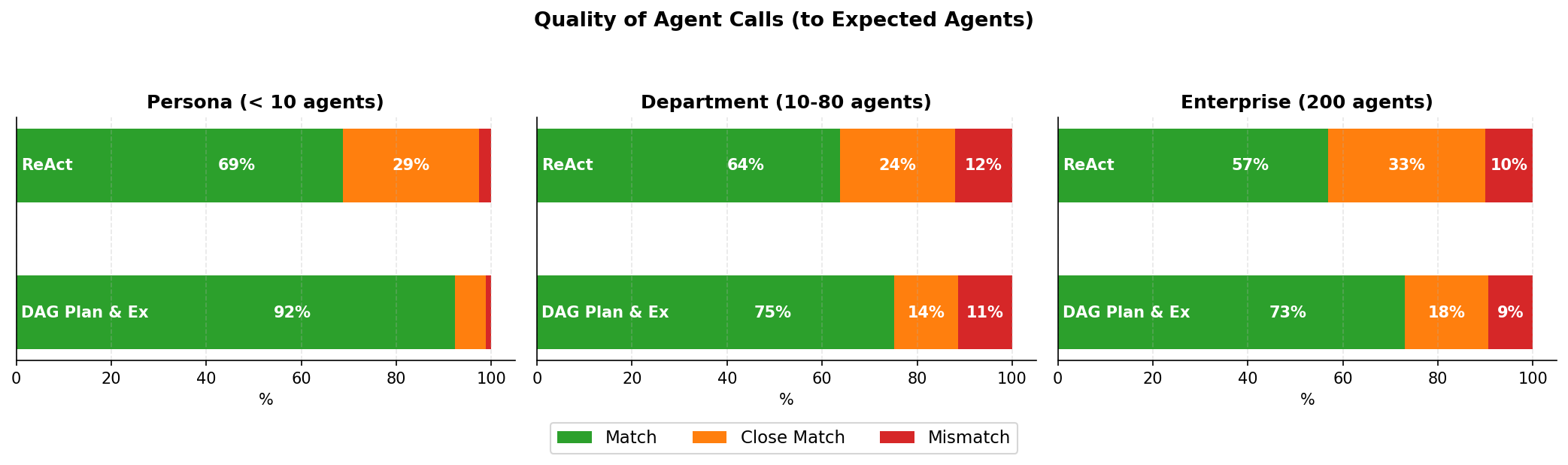}
\caption{Entity classification quality for responses from expected agents (Orchestration comparison).}\label{fig:classification-breakdown}
\end{figure}

\textbf{Quality of agent calls (Figure~\ref{fig:classification-breakdown}).} Figure~\ref{fig:classification-breakdown} evaluates entity-identification classification for responses from expected agents: Match, Close Match, and Mismatch. Exact Match is higher for DAG than ReAct in all three cohorts: 92\% vs 69\% at Persona, 75\% vs 64\% at Department, and 73\% vs 57\% at Enterprise. ReAct has the larger Close Match share in each cohort, while Mismatch remains between approximately 1\% and 12\%. This is an observed prompt-classification pattern; the benchmark does not isolate which system component produces it.

\begin{figure}[!tp]
\centering
\includegraphics[alt={Expected Answer Completeness heatmaps by business domain, scenario type, orchestrator, and Persona or Department cohort.},width=\linewidth,height=0.78\textheight,keepaspectratio]{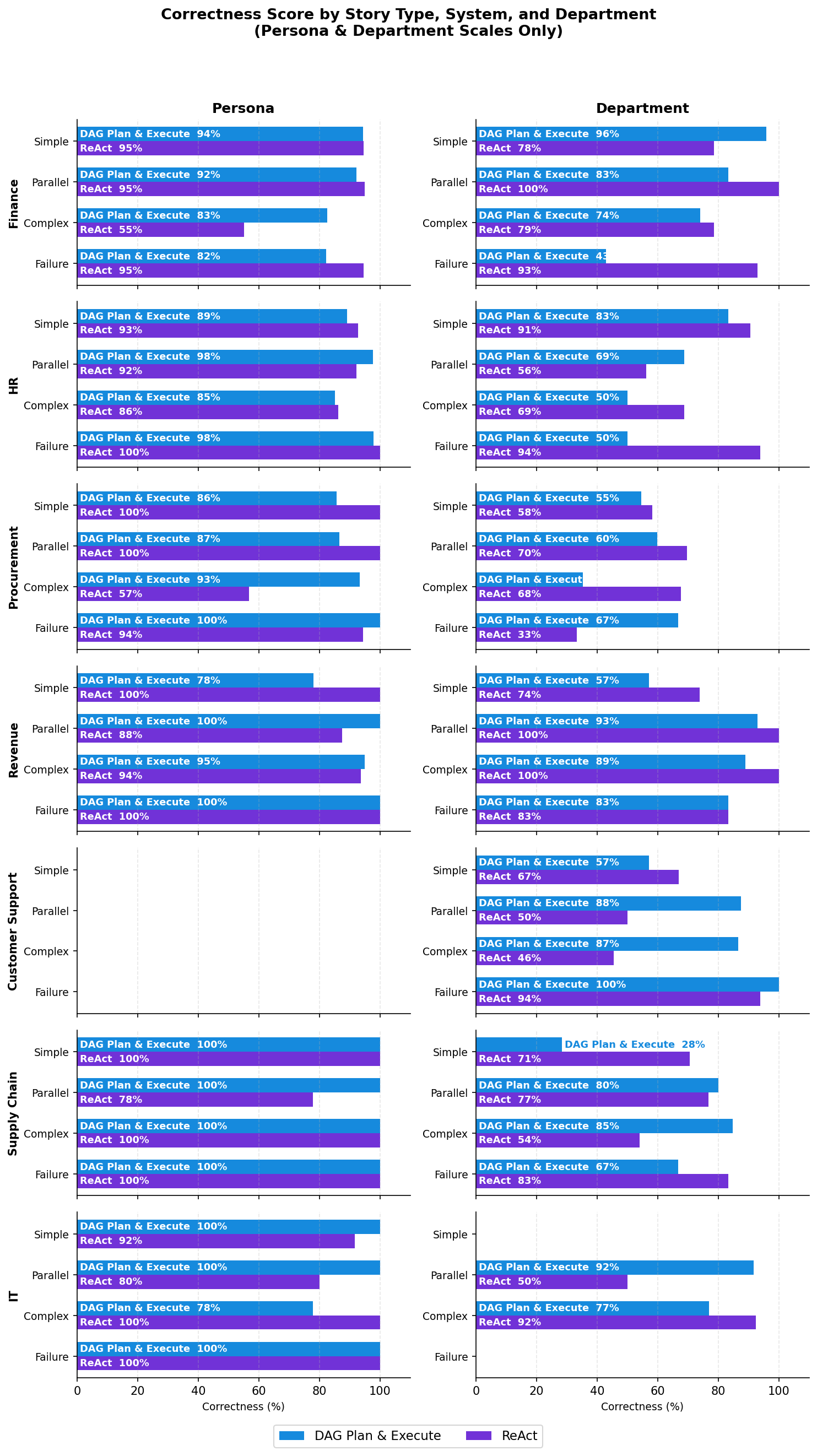}
\caption{Correctness scores by story type and business domain for DAG Plan \& Execute vs ReAct (Persona and Department scales).}\label{fig:correctness-domain}
\end{figure}

\textbf{Expected Answer Completeness by business domain (Figure~\ref{fig:correctness-domain}).} Figure~\ref{fig:correctness-domain} disaggregates the metric labeled \emph{Correctness} in the figure across business domains at Persona and Department scales. Scores vary widely by domain and scenario type, and the direction of the DAG--ReAct difference is not uniform. Some cells are absent, including Customer Support at Persona and Technology Platform Department Simple and Failure. The available cells therefore support a heterogeneity finding, not a consistent architecture advantage across business domains.

\FloatBarrier
\section{LLM-as-a-Judge Evaluation Prompt}\label{app:judge-prompt}

Response quality is evaluated using an LLM-as-a-Judge approach. Given the orchestrator's final synthesized response and a list of expected answer snippets (derived from agent guidelines' \texttt{expected\_response\_info} fields), the judge determines semantic presence of each expected snippet. The evaluation uses Claude Haiku for cost efficiency across 1,051 agent calls.

\textbf{Correctness Evaluation Prompt:}

\begin{lstlisting}
You are an evaluation assistant. Check if expected information snippets
are present in a response.

For EACH expected snippet, output `1` if present (exact or semantic match),
`0` if not present.

**Final Response:**
{final_response}

**Expected Answer Snippets:**
{expected_snippets_numbered}

Return JSON:
{
  "matches": [1, 0, 1],
  "reasoning": "Brief explanation per item"
}
\end{lstlisting}

The judge returns a binary match vector aligned with the expected snippets list. The Expected Answer Completeness score for each event is computed as:

$$\mathrm{EAC} = \frac{\sum_{i=1}^{n} \text{matches}[i]}{n} \times 100$$

where $n$ is the number of expected snippets. The figures and stored evaluation outputs use the shorter label \emph{Correctness}. The metric measures expected-fragment recall and does not independently penalize hallucinated, contradictory, irrelevant, or unsupported content.

\FloatBarrier
\section{DAG Planner Prompts}\label{app:dag-prompts}

The DAG Planner operates in two phases: agent discovery (selecting which agents to invoke) and DAG creation (constructing the execution graph). Both phases use Claude Sonnet 4.6 as the primary model.

\subsection{Phase 1: Agent Discovery}

The agent discovery prompt implements a two-step directed-reasoning approach: the LLM first articulates what information is needed before selecting agents. This separation reduces domain-label confusion when agent names are ambiguous.

\textbf{System Prompt:}

\begin{lstlisting}
You are an agent router. Given a task and available agents, select which
agents to call.

Steps:
1. List what information is needed (bullet points with '- ' prefix)
2. Match those needs to agents by their skills/descriptions

Rules:
- Select ALL agents that match any information need
  (comprehensive > minimal)
- If unsure whether an agent is relevant, include it -- false positives
  are cheap, false negatives lose information
- Use EXACT agent names from the list
- Return empty list only if no agent matches at all

Respond with JSON:
{"information_needed": "- item1\n- item2",
 "selected_agents": ["exact name", ...]}
\end{lstlisting}

\textbf{User Prompt Template:}

\begin{lstlisting}
Task: {task}

Available Agents:
{agent_list}

Identify the information needed and select the agents.
\end{lstlisting}

The \texttt{agent\_list} contains lightweight summaries for all registered agents, including skill descriptions, tags, and example queries. During replanning, additional context about previously failed agents and their error messages is appended to the user prompt.

\subsection{Phase 2: DAG Creation}

Once agents are selected, the DAG Creator constructs an execution plan as a Directed Acyclic Graph. The prompt enforces dependency-minimization: edges are only added when a downstream agent \emph{cannot start} without a specific data value from upstream.

\textbf{System Prompt:}

\begin{lstlisting}
You are an expert manager who must solve a problem by delegating tasks
to several agents.
You must create a plan in the form of a DAG (Directed Acyclic Graph).

Follow the instructions below:
1) Look at the `information needed` section and identify which agents
   have access to that information.
    1.1) In case you are unhappy with the information provided section
         feel free to make edits.
    1.2) Not all agents list will be relevant to the task.
2) Identify the dependencies, where information from one agent is a
   necessary prerequisite for the next.
3) Create the plan's DAG by filling in the JSON Schema:
    3.1) You MUST use one node per agent, unless the tasks are:
        3.1.1) Fundamentally Different in nature & Independent
        3.1.2) Require several different prerequisites
    3.2) Always include specific entity identifiers to help the agents
         identify which systems to act on. These include:
        3.2.1) Equipment IDs
        3.2.2) Order Numbers
        3.2.3) Plant Codes
        3.2.4) Customer IDs
        3.2.5) etc.
    3.3) Use DESCRIPTIVE node IDs like "fetch_equipment_telemetry"
         not "node_1"
    3.4) Only add dependencies (`depends_on`) when the downstream agent
         CANNOT START without a specific data value produced by the
         upstream agent.
        3.4.1) e.g., The agent needs an ID, amount, or classification
                that ONLY exists in the upstream output
        3.4.2) DO NOT depend just because upstream context would be
                "nice to have" or would "enrich" the result.
4) You MUST respond in the JSON Schema below:

### Response Format
{
  "$defs": {
    "SimplifiedNode": {
      "properties": {
        "agent": {
          "description": "Exact agent name from list",
          "type": "string"
        },
        "task": {
          "description": "What the agent should accomplish",
          "type": "string"
        },
        "information_required": {
          "description": "What data is needed from predecessors",
          "type": "string or null"
        },
        "constraints": {
          "description": "Filters, thresholds, or conditions",
          "type": "string or null"
        },
        "expected_output": {
          "description": "What information the agent should produce",
          "type": "string or null"
        },
        "depends_on": {
          "description": "Node IDs whose output this node needs",
          "type": "array of strings or null"
        }
      },
      "required": ["agent", "task"]
    }
  },
  "properties": {
    "reasoning": {
      "description": "Your reasoning for creating this plan",
      "type": "string or null"
    },
    "nodes": {
      "additionalProperties": { "$ref": "#/$defs/SimplifiedNode" },
      "type": "object"
    }
  },
  "required": ["reasoning", "nodes"]
}

Remember be clear, precise & purposeful. You MUST respond in the JSON
format given.
\end{lstlisting}

\textbf{User Prompt Template:}

\begin{lstlisting}
Create an execution plan for this task:
{task}

<information-needed>
    {information_needed}
</information-needed>

<available-agents>
    {agent_info}
</available-agents>

You MUST respond in the given JSON format.
\end{lstlisting}

The \texttt{information\_needed} field is passed directly from Phase 1's output, creating a chain-of-thought link between discovery and planning. The \texttt{agent\_info} section contains full agent cards (skills, descriptions, example queries) for only the selected agents.

\FloatBarrier
\section{ReAct Agent Prompt}\label{app:react-prompt}

The ReAct baseline (used by both the PydanticAI and TMPlus system variants) employs a single system prompt with tool-calling capabilities. Unlike the DAG Plan \& Execute architecture which separates planning from execution, the ReAct agent performs both in a single reasoning loop: it discovers agents, decides which to call, invokes them (potentially in parallel via native tool-calling), and synthesizes responses iteratively.

\textbf{System Prompt:}

\begin{lstlisting}
You are a multi-agent orchestration coordinator. Your job is to:

1. Analyze the incoming task/event description
2. Use `list_available_agents` to discover which specialized agents
   can help
3. Determine which agents to call and what to ask each one
4. Use `call_agent` to invoke agents -- you may call MULTIPLE agents
   simultaneously when their tasks are independent (the framework will
   execute them in parallel)
5. Synthesize all agent responses into a coherent final answer

Strategy guidelines:
- Be specific in your requests to each agent
- If an agent fails, note it and proceed with available information
- Synthesize results into a clear, actionable summary

Agent intelligence:
- Agents are intelligent and can handle complex, multi-part requests
  in a single call
- PREFER ONE CALL PER AGENT with a comprehensive request rather than
  splitting into multiple calls

You have access to external specialized agents via the A2A protocol.
Each agent has specific capabilities described in its card.
\end{lstlisting}

\textbf{Available Tools:}

\begin{itemize}[leftmargin=*,itemsep=0.1em,topsep=0.2em,parsep=0pt]
\item \texttt{list\_available\_agents}: returns all registered agent cards with names, descriptions, and skill summaries.
\item \texttt{call\_agent(agent\_name, message)}: invokes a specific agent via A2A; multiple concurrent calls are executed in parallel by the framework.
\end{itemize}

The ReAct architecture relies entirely on the LLM's reasoning capability to determine execution order, parallelism opportunities, and failure handling -- capabilities that the DAG Plan \& Execute architecture delegates to explicit algorithmic components (the DAG structure for ordering, \texttt{asyncio.gather} for parallelism, and the Node Outcome Decision prompt for failures). This is the central architectural contrast, but the tested systems also differ in discovery timing, planning, execution, and recovery policy; the evaluation therefore does not isolate a single architectural independent variable.

\FloatBarrier
\clearpage
\begingroup
\RaggedRight
\section{Agent Catalog}\label{app:agent-catalog}

The evaluation employs a catalog of 200 enterprise agents spanning seven business domains: Finance (69), HR (47), Procurement (24), Revenue (21), Customer Support (11), Supply Chain (9), and Technology Platform (8). Agents are grouped under personas -- user-facing categories that organize related capabilities.

\subsection*{I. Finance}
\subsubsection*{I-A. Financial Close Operator}
\noindent\hangindent=2em\hangafter=1\makebox[2em][l]{\textbf{001.}}\textbf{Accruals}.~Automates journal entry preparation for manual accruals, ensuring completeness and accuracy in the financial close. Calculates required accrual amounts based on business rules, historical patterns, and period-end cutoff requirements.\par
\noindent\hangindent=2em\hangafter=1\makebox[2em][l]{\textbf{002.}}\textbf{Analytical Insights}.~Assesses financial performance through KPIs and generates concise summary reports for stakeholders. Provides real-time visibility into close progress, variance analysis, and key financial metrics to support decision-making during the close process.\par
\noindent\hangindent=2em\hangafter=1\makebox[2em][l]{\textbf{003.}}\textbf{Anomaly Detection}.~Identifies unusual patterns and anomalies in financial close data to catch errors before they propagate, enabling 90\% faster resolution of exceptions. Monitors transaction flows, account balances, and posting patterns against expected behavior.\par
\noindent\hangindent=2em\hangafter=1\makebox[2em][l]{\textbf{004.}}\textbf{AR AP Clearing}.~Clears open items via agent-generated matching proposals, reducing manual matching effort by 80\%. Uses intelligent algorithms to match payments, invoices, and credit notes across AP and AR, handling complex scenarios like partial payments and multiple invoice settlements.\par
\noindent\hangindent=2em\hangafter=1\makebox[2em][l]{\textbf{005.}}\textbf{Asset Accounting}.~Manages fixed asset lifecycle events, depreciation runs, and period-end asset postings.\par
\noindent\hangindent=2em\hangafter=1\makebox[2em][l]{\textbf{006.}}\textbf{Error Resolution}.~Resolves errors in production cost postings and other financial close transactions with AI-powered root cause analysis. Achieves 90\% faster resolution by automatically diagnosing issues and proposing corrective actions.\par
\noindent\hangindent=2em\hangafter=1\makebox[2em][l]{\textbf{007.}}\textbf{Financial Consistency Agent}.~Analyzes financial data inconsistencies based on errors received by end users, evaluates them using the FDCA (Financial Data Consistency Analysis) framework and historical patterns, and proposes resolutions wherever possible to accelerate period-end closing and ensure data integrity.\par
\noindent\hangindent=2em\hangafter=1\makebox[2em][l]{\textbf{008.}}\textbf{GL Account Review}.~Reviews GL account balances for completeness and accuracy during close.\par
\noindent\hangindent=2em\hangafter=1\makebox[2em][l]{\textbf{009.}}\textbf{GL Clearing}.~Clears open items on general ledger accounts based on matching rules.\par
\noindent\hangindent=2em\hangafter=1\makebox[2em][l]{\textbf{010.}}\textbf{GR IR Clearing}.~Resolves goods receipt/invoice receipt discrepancies for period-end close.\par
\noindent\hangindent=2em\hangafter=1\makebox[2em][l]{\textbf{011.}}\textbf{Intercompany}.~Automates matching of intercompany invoices and cross charges using embedded AI, eliminating manual reconciliation effort. Identifies discrepancies between intercompany trading partners and proposes resolution actions to accelerate close.\par
\noindent\hangindent=2em\hangafter=1\makebox[2em][l]{\textbf{012.}}\textbf{Journal Entry}.~Automates preparation of journal entries from external data sources, significantly reducing manual effort in the financial close process. Handles recurring entries, accruals, and adjustments with intelligent data extraction and validation to achieve 70\% faster zero-touch journal entry processing.\par
\noindent\hangindent=2em\hangafter=1\makebox[2em][l]{\textbf{013.}}\textbf{Reconciliation Agent Extended Intercompany Scope}.~Automates matching of intercompany invoices and cross charges using embedded AI, eliminating manual reconciliation effort. Identifies discrepancies between intercompany trading partners and proposes resolution actions to accelerate close.\par

\subsubsection*{I-B. Accounts Receivable Manager}
\noindent\hangindent=2em\hangafter=1\makebox[2em][l]{\textbf{014.}}\textbf{Aging Analysis}.~Analyzes receivable aging buckets with predictive insights on payment timing.\par
\noindent\hangindent=2em\hangafter=1\makebox[2em][l]{\textbf{015.}}\textbf{AR Account Analysis}.~Provides deep analysis of customer accounts for patterns, risks, and optimization opportunities. Processes data related to overdue receivables from various applications and performs appropriate follow-up tasks with customers based on their profile and payment history.\par
\noindent\hangindent=2em\hangafter=1\makebox[2em][l]{\textbf{016.}}\textbf{AR Worklist Priority}.~Intelligently prioritizes accounts receivable work items using AI-driven analysis of payment likelihood, customer risk profiles, and revenue impact to ensure AR teams focus on the highest-value activities first.\par
\noindent\hangindent=2em\hangafter=1\makebox[2em][l]{\textbf{017.}}\textbf{Collection 360}.~Provides model-driven predictive prioritization and outreach for collections, combining customer payment patterns, risk scores, and communication history to optimize collection strategies and reduce days sales outstanding.\par
\noindent\hangindent=2em\hangafter=1\makebox[2em][l]{\textbf{018.}}\textbf{Credit Risk Analysis}.~Monitors internal and external signals to assess customer credit risk in real-time, dynamically adjusting credit limits based on financial health indicators, payment behavior, and market conditions. Enables proactive credit management to minimize bad debt exposure.\par
\noindent\hangindent=2em\hangafter=1\makebox[2em][l]{\textbf{019.}}\textbf{Customer Contact Management}.~Manages customer communication history and schedules follow-ups for collections.\par
\noindent\hangindent=2em\hangafter=1\makebox[2em][l]{\textbf{020.}}\textbf{Dispute Management}.~Correlates deduction claims with service and delivery data to efficiently manage and resolve AR disputes. Automates research, validation, and resolution workflows to shorten dispute cycle time and protect revenue.\par
\noindent\hangindent=2em\hangafter=1\makebox[2em][l]{\textbf{021.}}\textbf{Dunning Agent}.~Automates the dunning process by generating and sending contextually appropriate payment reminders based on customer segments, payment behavior, and relationship status to maximize collection effectiveness.\par
\noindent\hangindent=2em\hangafter=1\makebox[2em][l]{\textbf{022.}}\textbf{Promise To Pay}.~Tracks and manages customer payment commitments, escalating broken promises automatically.\par
\noindent\hangindent=2em\hangafter=1\makebox[2em][l]{\textbf{023.}}\textbf{Resubmission Agent}.~Automatically resubmits failed or rejected items at optimal intervals.\par
\noindent\hangindent=2em\hangafter=1\makebox[2em][l]{\textbf{024.}}\textbf{Smart Clearing}.~Achieves 95\%+ auto-match rates using AI pattern recognition to automatically clear incoming payments against open invoices, reducing manual effort and accelerating cash application across high-volume transactions.\par

\subsubsection*{I-C. Accounts Payable Manager}
\noindent\hangindent=2em\hangafter=1\makebox[2em][l]{\textbf{025.}}\textbf{AP Aging}.~Analyzes payable aging with cash flow optimization recommendations.\par
\noindent\hangindent=2em\hangafter=1\makebox[2em][l]{\textbf{026.}}\textbf{Discount Optimization}.~Identifies and captures early payment discount opportunities across vendor base.\par
\noindent\hangindent=2em\hangafter=1\makebox[2em][l]{\textbf{027.}}\textbf{Duplicate Detection}.~Identifies potential duplicate invoices using fuzzy matching across multiple data points.\par
\noindent\hangindent=2em\hangafter=1\makebox[2em][l]{\textbf{028.}}\textbf{Invoice Capture}.~Extracts information from invoices sent as PDF (or images) leveraging multiple AI engines -- M/L and LLMs -- to achieve highest levels of accuracy.\par
\noindent\hangindent=2em\hangafter=1\makebox[2em][l]{\textbf{029.}}\textbf{Invoice Validation}.~Validates invoice data against purchase orders, contracts, and business rules.\par
\noindent\hangindent=2em\hangafter=1\makebox[2em][l]{\textbf{030.}}\textbf{Payment Execution}.~Executes payments with intelligent timing to maximize cash position and early payment discounts.\par
\noindent\hangindent=2em\hangafter=1\makebox[2em][l]{\textbf{031.}}\textbf{Payment Proposal}.~Generates optimized payment proposals considering cash flow, discounts, and payment terms.\par
\noindent\hangindent=2em\hangafter=1\makebox[2em][l]{\textbf{032.}}\textbf{Vendor Reconciliation}.~Reconciles vendor statements against AP records, identifying and resolving discrepancies.\par

\subsubsection*{I-D. Billing \& Invoicing Specialist}
\noindent\hangindent=2em\hangafter=1\makebox[2em][l]{\textbf{033.}}\textbf{Billing Adjustment}.~Provides accurate, fast correction of billing errors to reduce revenue leakage from rejected or disputed invoices. Handles credit/debit memos, retroactive price changes, and billing corrections with full audit trail to maintain financial integrity.\par
\noindent\hangindent=2em\hangafter=1\makebox[2em][l]{\textbf{034.}}\textbf{Billing Anomaly Detection}.~Catches billing anomalies early to reduce rework, disputes, and compliance risks. Monitors billing patterns to detect duplicate charges, unusual amounts, timing irregularities, and deviations from expected billing behavior before they become costly errors.\par
\noindent\hangindent=2em\hangafter=1\makebox[2em][l]{\textbf{035.}}\textbf{Billing Creation}.~Automates the prioritization and creation of billing documents from fulfilled orders, applying correct pricing, tax determination, and customer-specific policy rules. Reduces manual effort in invoice creation by intelligently processing billing due list items.\par
\noindent\hangindent=2em\hangafter=1\makebox[2em][l]{\textbf{036.}}\textbf{Billing Due List}.~Automates the prioritization of billing due list items using customer policy documents. Identifies and ranks pending billing items based on contract terms, delivery milestones, and revenue recognition rules to ensure timely and accurate billing execution.\par
\noindent\hangindent=2em\hangafter=1\makebox[2em][l]{\textbf{037.}}\textbf{Billing Posting}.~Provides AI-powered explanations and resolution proposals that massively accelerate posting error handling. Automatically generates billing documents from fulfilled orders and posts them with correct account determination, ensuring smooth financial close.\par
\noindent\hangindent=2em\hangafter=1\makebox[2em][l]{\textbf{038.}}\textbf{Dispute Resolution}.~Manages end-to-end billing dispute resolution by analyzing dispute cases, identifying root causes, and proposing corrective actions. Accelerates the resolution cycle by leveraging historical patterns and customer communication history.\par

\subsubsection*{I-E. Travel \& Expense Manager}
\noindent\hangindent=2em\hangafter=1\makebox[2em][l]{\textbf{039.}}\textbf{Booking Agent}.~Powered by an enterprise AI assistant, the booking agent in a travel management platform delivers personalized flight and hotel recommendations by analyzing individual traveler preferences, company travel policies, and budget constraints. Streamlines the booking process, reduces time spent on manual search, and increases compliance with company guidelines.\par
\noindent\hangindent=2em\hangafter=1\makebox[2em][l]{\textbf{040.}}\textbf{Expense Pre Submit Audit Agent}.~For expense report submitters, the Expense Pre-Submit Audit Agent automatically checks expenses for receipt accuracy issues and policy breaches before submission. The result is fewer report rejections, faster reimbursements, and a smoother employee experience.\par
\noindent\hangindent=2em\hangafter=1\makebox[2em][l]{\textbf{041.}}\textbf{Expense Report Validation Agent}.~Validates expense reports for policy compliance, detects anomalies, and ensures accurate categorization before submission. Leverages AI to check expense claims against company policies, flag potential issues, and recommend corrections to reduce rejection rates and speed up approval cycles.\par
\noindent\hangindent=2em\hangafter=1\makebox[2em][l]{\textbf{042.}}\textbf{Expense Receipt Analysis Agent}.~The receipt analysis agent leverages multiple tools, including maps, vendor databases, web searches, and trip itineraries from a travel management platform, to itemize and categorize receipt data for business expense submitters. It reasons over data on the receipt and contextual data outside the receipt, such as map data and business trip itineraries, to accurately create expense entries.\par
\noindent\hangindent=2em\hangafter=1\makebox[2em][l]{\textbf{043.}}\textbf{Meeting Location Planner Agent}.~Suggests offsite meeting locations that minimize attendee travel time, account for office proximity, identify suitable hotels, estimate total trip cost, and prepare booking communication for attendees.\par

\subsubsection*{I-F. Financial Planning \& Analysis}
\noindent\hangindent=2em\hangafter=1\makebox[2em][l]{\textbf{044.}}\textbf{Budgeting Forecasting}.~AI-driven budget creation and rolling forecasts with automatic variance detection.\par
\noindent\hangindent=2em\hangafter=1\makebox[2em][l]{\textbf{045.}}\textbf{Outlier Analysis}.~Detects statistical outliers in financial data requiring investigation.\par
\noindent\hangindent=2em\hangafter=1\makebox[2em][l]{\textbf{046.}}\textbf{Risk Analysis Treasury Assistant}.~Assesses financial risks across the enterprise with probability-weighted impact.\par
\noindent\hangindent=2em\hangafter=1\makebox[2em][l]{\textbf{047.}}\textbf{Scenario Planning}.~Generates and evaluates multiple financial scenarios with impact analysis.\par
\noindent\hangindent=2em\hangafter=1\makebox[2em][l]{\textbf{048.}}\textbf{Variance Analysis}.~Automatically identifies and explains significant variances between actuals and plan.\par
\noindent\hangindent=2em\hangafter=1\makebox[2em][l]{\textbf{049.}}\textbf{Risk Analysis Financial Planning Assistant}.~Assesses financial risks across the enterprise with probability-weighted impact.\par

\subsubsection*{I-G. Purchase Order Execution}
\noindent\hangindent=2em\hangafter=1\makebox[2em][l]{\textbf{050.}}\textbf{Buying Agent}.~Automates and optimizes the procurement process for goods and services by identifying suitable suppliers, comparing quotes, evaluating vendor proposals, and recommending optimal purchasing decisions based on price, quality, delivery time, and compliance requirements. Streamlines buy decisions and reduces procurement cycle time.\par

\subsubsection*{I-H. Cash Collections Specialist}
\noindent\hangindent=2em\hangafter=1\makebox[2em][l]{\textbf{051.}}\textbf{Case Preperation Agent}.~Prepares comprehensive case documentation and analysis for utilities customer service and billing disputes by gathering relevant account data, transaction history, usage patterns, and policy information. Organizes evidence and context needed for case resolution, reducing preparation time and improving case outcome quality.\par
\noindent\hangindent=2em\hangafter=1\makebox[2em][l]{\textbf{052.}}\textbf{Case Processing Agent Cash Collections Assistant}.~Processes utilities customer service cases by analyzing issues, applying business rules, identifying resolution paths, and executing approved actions to resolve billing disputes, service requests, and customer inquiries. Accelerates case resolution while maintaining accuracy and customer satisfaction.\par
\noindent\hangindent=2em\hangafter=1\makebox[2em][l]{\textbf{053.}}\textbf{Collection Agent}.~Manages utility collections processes by prioritizing overdue accounts, determining optimal collection strategies, generating payment reminders, and tracking collection activities. Helps utilities improve cash flow and reduce delinquencies while maintaining positive customer relationships.\par
\noindent\hangindent=2em\hangafter=1\makebox[2em][l]{\textbf{054.}}\textbf{Dispute Resolution Agent}.~This agent automates root-cause analysis and suggests actionable solutions. It quickly reviews invoices, sales orders, delivery records, pricing agreements, tax rules and historical data to pinpoint dispute causes. Offering compliant, tailored recommendations, it cuts resolution time and effort by analyzing large datasets, boosting efficiency and satisfaction. It streamlines workflows, reduces resource use, and advances automated dispute management for smarter, faster, regulatory-aligned resolutions.\par
\noindent\hangindent=2em\hangafter=1\makebox[2em][l]{\textbf{055.}}\textbf{Invoicing Agent}.~Automates utility invoice generation and delivery by consolidating usage data, applying correct rates and tariffs, calculating charges, and producing accurate bills for customers. Ensures timely billing cycles, proper charge categorization, and integration with payment systems for utilities operations.\par
\noindent\hangindent=2em\hangafter=1\makebox[2em][l]{\textbf{056.}}\textbf{Payment Matching Agent}.~Matches incoming utility payments to outstanding invoices and accounts using intelligent algorithms that handle partial payments, multiple invoice settlements, and complex payment scenarios. Reduces manual reconciliation effort and accelerates cash application for utilities billing operations.\par

\subsubsection*{I-I. Treasury \& Cash Management}
\noindent\hangindent=2em\hangafter=1\makebox[2em][l]{\textbf{057.}}\textbf{Cash Insights}.~Provides real-time cash position visibility and predictive cash flow analytics.\par
\noindent\hangindent=2em\hangafter=1\makebox[2em][l]{\textbf{058.}}\textbf{Working Capital Agent}.~Optimizes working capital through intelligent management of receivables, payables, and inventory.\par

\subsubsection*{I-J. Financial Reporting Analyst}
\noindent\hangindent=2em\hangafter=1\makebox[2em][l]{\textbf{059.}}\textbf{Consolidation}.~Automates group consolidation including currency translation and elimination entries.\par
\noindent\hangindent=2em\hangafter=1\makebox[2em][l]{\textbf{060.}}\textbf{Public Disclosure Narrative}.~Generates compliant disclosure narratives for financial reporting.\par

\subsubsection*{I-K. Document \& Reporting Compliance}
\noindent\hangindent=2em\hangafter=1\makebox[2em][l]{\textbf{061.}}\textbf{E Invoicing Setup Agent}.~The e-invoicing setup agent takes over from legal-change discovery and guides the implementation lead with a detailed explanation of e-invoicing mandates and impacted business scenarios. It reduces manual setup across ERP and compliance applications by outlining a guided workflow and proposing relevant configuration values for human review, helping teams reach mandate readiness on time.\par
\noindent\hangindent=2em\hangafter=1\makebox[2em][l]{\textbf{062.}}\textbf{Legal Change Agent}.~The legal-change agent enables customers to discover upcoming regulatory changes affecting their enterprise ERP system. It provides comprehensive information on each regulation, including a concise legal summary, affected enterprise products, compliance deadlines, and solution summary, helping customers plan and maintain continuity of business operations.\par
\noindent\hangindent=2em\hangafter=1\makebox[2em][l]{\textbf{063.}}\textbf{Localization Extensibility Agent}.~The localization extensibility agent automates the merge of customer-extended objects in statutory reports in an enterprise compliance application. When new report versions are delivered, the agent proactively detects impacted extensions via system events, analyzes field mappings and conflicts, and generates a merged draft extension for user review or autonomous activation through natural language. It reduces extension migration time, lowers error rates through AI-validated merging, and minimizes compliance risk across report types and countries.\par

\subsubsection*{I-L. Contingent Workforce Coordinator}
\noindent\hangindent=2em\hangafter=1\makebox[2em][l]{\textbf{064.}}\textbf{Job Creation Agent}.~Streamlines the creation of job requisitions and postings by automatically generating comprehensive job descriptions, determining appropriate compensation ranges, identifying required skills and qualifications, and publishing to relevant job boards. Reduces time-to-post and ensures consistent, compliant job listings.\par
\noindent\hangindent=2em\hangafter=1\makebox[2em][l]{\textbf{065.}}\textbf{Rate Guidance Agent}.~Provides intelligent guidance on contractor and contingent worker rates by analyzing market data, historical rates, role requirements, location factors, and skill demand to recommend competitive yet budget-conscious pricing. Helps procurement and HR teams negotiate fair rates and maintain cost efficiency.\par
\noindent\hangindent=2em\hangafter=1\makebox[2em][l]{\textbf{066.}}\textbf{Review Assessment Agent}.~Conducts comprehensive reviews and assessments of contractor performance, project deliverables, and service quality to support renewal decisions, rate adjustments, and vendor management. Analyzes multiple data points including performance metrics, feedback, and historical patterns to provide actionable recommendations.\par

\subsubsection*{I-M. Workforce Management Specialist}
\noindent\hangindent=2em\hangafter=1\makebox[2em][l]{\textbf{067.}}\textbf{Mobilization Agent}.~Helps teams turn approved work orders into start-ready assignments by assembling role and location-based onboarding checklists, checking rate/tenure/right-to-work requirements, capturing and verifying required documents, highlighting risks, and routing exceptions for quick approval so workers start on time and in compliance.\par
\noindent\hangindent=2em\hangafter=1\makebox[2em][l]{\textbf{068.}}\textbf{SOW Creation Agent}.~Automates the creation of Statements of Work (SOW) for professional services engagements by generating comprehensive scope documents that include project objectives, deliverables, timelines, resource requirements, and pricing. Streamlines SOW creation process and ensures consistency across service contracts.\par

\subsubsection*{I-N. Governance \& Risk Oversight}
\noindent\hangindent=2em\hangafter=1\makebox[2em][l]{\textbf{069.}}\textbf{Regulatory Compliance}.~Tracks regulatory changes and assesses impact on financial processes.\par
\noindent\hangindent=2em\hangafter=1\makebox[2em][l]{\textbf{070.}}\textbf{Risk Assurance Management}.~Monitors control effectiveness and manages risk assurance activities.\par

\subsection*{II. HR}
\subsubsection*{II-A. Learning \& Development Coordinator}
\noindent\hangindent=2em\hangafter=1\makebox[2em][l]{\textbf{071.}}\textbf{Admin Assignment Agent}.~Aligns learning investments to actual business strategy, identifying where content gaps are blocking organizational goals. Ensures learning resources are strategically allocated to support business priorities.\par
\noindent\hangindent=2em\hangafter=1\makebox[2em][l]{\textbf{072.}}\textbf{Course Tutor Agent}.~Keeps learning live and continuous by managing different forms of consumption that meet users' needs, making knowledge accessible and retained. Provides contextual learning support during courses and activities.\par
\noindent\hangindent=2em\hangafter=1\makebox[2em][l]{\textbf{073.}}\textbf{Learning Compliance Agent}.~Monitors regulations, tracks individual compliance status, auto-enrolls users in required training, manages prerequisites, and sends proactive alerts. Ensures built-in guardrails prevent compliance failures and protect the organization.\par
\noindent\hangindent=2em\hangafter=1\makebox[2em][l]{\textbf{074.}}\textbf{Learning QA Skill}.~Provides instant answers to learning and development questions, helping employees and managers navigate the learning ecosystem. Reduces time spent searching for information about available programs, policies, and requirements.\par
\noindent\hangindent=2em\hangafter=1\makebox[2em][l]{\textbf{075.}}\textbf{Learning Recommendation Agent}.~Creates personalized, skills-based learning pathways that adapt in real-time to individual progress and organizational needs. Achieves 2-3x higher completion rates by recommending content that feels relevant and engaging.\par
\noindent\hangindent=2em\hangafter=1\makebox[2em][l]{\textbf{076.}}\textbf{Manager Assignment Agent}.~Combines skills, mobility, readiness, succession, performance and learning signals with financial models to predict and manage the ROI of talent investments. Helps managers make data-driven learning assignment decisions.\par
\noindent\hangindent=2em\hangafter=1\makebox[2em][l]{\textbf{077.}}\textbf{Manager Skill Gap Agent}.~Analyzes skill gaps across a manager's team and recommends targeted learning interventions to close critical gaps. Provides visibility into team readiness and development priorities.\par

\subsubsection*{II-B. Talent Acquisition Specialist}
\noindent\hangindent=2em\hangafter=1\makebox[2em][l]{\textbf{078.}}\textbf{Agent Connect}.~Enables seamless coordination between specialized recruiting agents, acting as an orchestration layer that routes requests to the appropriate agent and combines results for comprehensive responses to complex recruiting queries.\par
\noindent\hangindent=2em\hangafter=1\makebox[2em][l]{\textbf{079.}}\textbf{Candidate Relationship Management Agent}.~Automates discovery and engagement, transforming static customer relationship management data into proactive recruitment and reducing recruiter manual hours per hire by 30-50\%. Maintains candidate relationships through intelligent nurturing campaigns.\par
\noindent\hangindent=2em\hangafter=1\makebox[2em][l]{\textbf{080.}}\textbf{Interview Agent}.~AI-led interviews offer on-demand structured voice or video sessions for faster high-volume screening. Conducts consistent, unbiased initial interviews at scale, freeing recruiter time for high-value candidate interactions.\par
\noindent\hangindent=2em\hangafter=1\makebox[2em][l]{\textbf{081.}}\textbf{Interview Scheduling Agent}.~Automates interview coordination, reducing average time-to-schedule by 20\% and saving hours of manual effort per team. Handles complex multi-panel scheduling, timezone management, and rescheduling without human intervention.\par
\noindent\hangindent=2em\hangafter=1\makebox[2em][l]{\textbf{082.}}\textbf{Recruiting Migration Agents}.~Automates data validation and mapping between an HR management platform and a recruiting platform, significantly reducing implementation risk and time-to-go-live by 40\%. Handles complex data transformations and ensures consistency during system transitions.\par
\noindent\hangindent=2em\hangafter=1\makebox[2em][l]{\textbf{083.}}\textbf{Recruiting QA Agent}.~Accelerates workflow execution and self-service resolution by delivering instant, context-aware answers and seamless cross-system coordination. Acts as the primary interface for recruiter questions about processes, policies, and candidate status.\par
\noindent\hangindent=2em\hangafter=1\makebox[2em][l]{\textbf{084.}}\textbf{Recruiting QA Skill}.~Delivers instant, context-aware answers to recruiter and hiring-team queries by combining live recruiting-platform data with knowledge-base content. Enables users to get accurate guidance, resolve questions, and navigate processes within their workflow without needing to search documentation or switch systems. Builds trust by providing reliable, real-time assistance even when no direct action is required.\par
\noindent\hangindent=2em\hangafter=1\makebox[2em][l]{\textbf{085.}}\textbf{Reverse Matching Agent}.~Boosts pipeline quality and conversion rates by instantly connecting candidates to their best-fit roles. Analyzes candidate profiles against all open requisitions to proactively surface matches that recruiters might miss.\par

\subsubsection*{II-C. Career \& Talent Development}
\noindent\hangindent=2em\hangafter=1\makebox[2em][l]{\textbf{086.}}\textbf{Career Development Agent}.~Ensures personalized career guidance, increased employee satisfaction, talent readiness, and ultimately retention of critical talent. Provides individualized career path recommendations based on skills, aspirations, and organizational needs.\par
\noindent\hangindent=2em\hangafter=1\makebox[2em][l]{\textbf{087.}}\textbf{Succession Planning Agent}.~Increases succession health and talent readiness while reducing business risk for critical roles. Identifies succession gaps, evaluates bench strength, and recommends development actions to achieve 100\% critical role coverage.\par
\noindent\hangindent=2em\hangafter=1\makebox[2em][l]{\textbf{088.}}\textbf{Talent Scout Agent}.~Unleashes internal workforce potential by uncovering and matching talents to strategic work, roles, and tasks. Achieves 50\% efficiency gains in finding critical talent by analyzing skills, mobility preferences, and readiness across the organization.\par
\noindent\hangindent=2em\hangafter=1\makebox[2em][l]{\textbf{089.}}\textbf{Team Growth Agent}.~Creates personalized, skills-based learning pathways that adapt in real-time, moving beyond one-size-fits-all training that employees ignore. Combines career development with learning recommendations for holistic growth.\par

\subsubsection*{II-D. Time \& Attendance Tracker}
\noindent\hangindent=2em\hangafter=1\makebox[2em][l]{\textbf{090.}}\textbf{Configuration Agent}.~The routing configuration agent simplifies and automates document routing configuration for suppliers on a procurement network. By leveraging AI, LLMs, and a conversational interface, suppliers can configure routing using natural language instead of manual setup. Today, suppliers must manually configure routing rules for each document type, requiring understanding of formats such as cXML, EDI, and email delivery methods; this can lead to confusion, inconsistent configurations, and onboarding delays.\par
\noindent\hangindent=2em\hangafter=1\makebox[2em][l]{\textbf{091.}}\textbf{Timesheet Agent}.~Assists employees with timesheet entry, validation, and approval by auto-populating time entries based on calendar events, project assignments, and historical patterns. Validates entries against project codes and work rules, and streamlines approval workflows to reduce administrative burden.\par

\subsubsection*{II-E. System Administration \& Config}
\noindent\hangindent=2em\hangafter=1\makebox[2em][l]{\textbf{092.}}\textbf{Configuration Transport Center Agent}.~Automates and orchestrates configuration transport processes across system landscapes by managing transport requests, coordinating approvals, executing transports, and tracking deployment status. Reduces manual effort and errors in moving configurations between development, test, and production environments.\par
\noindent\hangindent=2em\hangafter=1\makebox[2em][l]{\textbf{093.}}\textbf{Role Based Permission Agent}.~Manages role-based permissions and access control by analyzing user roles, assigning appropriate permissions, detecting access conflicts, and ensuring compliance with security policies. Simplifies permission management and reduces security risks from excessive or inappropriate access.\par
\noindent\hangindent=2em\hangafter=1\makebox[2em][l]{\textbf{094.}}\textbf{Time Config Agent}.~The time configuration agent helps administrators by automatically proposing local public holidays sourced from authorized websites. Once approved, these proposals are created as configuration objects in the HR management platform. The goal is to reduce manual maintenance and ensure holiday data remains accurate and up to date.\par

\subsubsection*{II-F. Core HR Administrator}
\noindent\hangindent=2em\hangafter=1\makebox[2em][l]{\textbf{095.}}\textbf{Employee Data Integration Agent Core HR Assistant}.~Automates the resolution of data integration inconsistencies between the core HR system and downstream systems. Detects mismatches, proposes corrections, and ensures data consistency across the HR technology landscape.\par
\noindent\hangindent=2em\hangafter=1\makebox[2em][l]{\textbf{096.}}\textbf{Payment Maintenance Agent}.~Ensures payment information changes are valid, detecting potential fraud and verifying completeness. Validates bank account changes, routing information, and payment method updates against fraud patterns and organizational policies.\par
\noindent\hangindent=2em\hangafter=1\makebox[2em][l]{\textbf{097.}}\textbf{Pending Workflow Agent Time Assistant}.~Manages pending workflow items and approvals by prioritizing requests, routing to appropriate approvers, tracking approval status, identifying bottlenecks, and escalating overdue items. Accelerates decision-making and reduces workflow cycle times across the organization.\par
\noindent\hangindent=2em\hangafter=1\makebox[2em][l]{\textbf{098.}}\textbf{Person Data Quality Agent}.~Corrects and correlates data changes across objects to ensure zero compliance issues for person data. Monitors data quality metrics and proactively identifies records that need correction to maintain regulatory compliance.\par
\noindent\hangindent=2em\hangafter=1\makebox[2em][l]{\textbf{099.}}\textbf{Position Management Agent}.~Ensures managers use the right positions for new headcount asks and transfers, validating position availability, budget alignment, and organizational structure compliance. Streamlines the position lifecycle from creation to fill.\par
\noindent\hangindent=2em\hangafter=1\makebox[2em][l]{\textbf{100.}}\textbf{Stalled Workflow Agent}.~Accelerates stalled workflows by nudging, rerouting, approving or declining workflows that have been pending beyond acceptable thresholds. Reduces bottlenecks in HR processes and ensures timely completion of approvals.\par
\noindent\hangindent=2em\hangafter=1\makebox[2em][l]{\textbf{101.}}\textbf{Pending Workflow Agent Core HR Assistant}.~Identifies and manages pending workflows that require attention, ensuring timely processing of HR transactions. Works in conjunction with the Stalled Workflow Agent to maintain workflow velocity across all HR processes.\par

\subsubsection*{II-G. Performance \& Goal Management}
\noindent\hangindent=2em\hangafter=1\makebox[2em][l]{\textbf{102.}}\textbf{Feedback Catalyst Agent}.~Detects rating bias and ensures fair performance distributions across demographic groups, reducing compliance exposure while improving retention of historically underrated talent. Provides calibration insights to promote equity.\par
\noindent\hangindent=2em\hangafter=1\makebox[2em][l]{\textbf{103.}}\textbf{Goal Creation Agent}.~Generates strategically aligned, high-quality goals conversationally, helping employees and managers create meaningful objectives that connect individual work to organizational strategy. Increases likelihood of employees understanding their work's strategic connection by 2-3x.\par
\noindent\hangindent=2em\hangafter=1\makebox[2em][l]{\textbf{104.}}\textbf{Goal Progress Agent}.~Monitors goal health and surfaces risks proactively, ensuring managers and employees stay on track throughout the performance cycle. Provides early warning when goals are at risk and suggests corrective actions.\par
\noindent\hangindent=2em\hangafter=1\makebox[2em][l]{\textbf{105.}}\textbf{Performance Intelligence Agent}.~Continuously captures and formalizes evidence of contributions, creating a comprehensive performance record that reduces recency bias and supports fair evaluations. Surfaces 20-25\% more internal promotion-qualified candidates.\par
\noindent\hangindent=2em\hangafter=1\makebox[2em][l]{\textbf{106.}}\textbf{Performance Preparation Agent}.~Equips managers with data-driven insights for 1:1 conversations and performance reviews. Synthesizes goal progress, peer feedback, contribution evidence, and development areas into actionable briefings that improve review quality.\par

\subsubsection*{II-H. HR Advisory \& Policy}
\noindent\hangindent=2em\hangafter=1\makebox[2em][l]{\textbf{107.}}\textbf{G P GIA Agent}.~Helps HR teams stay compliant across 50 countries.\par
\noindent\hangindent=2em\hangafter=1\makebox[2em][l]{\textbf{108.}}\textbf{Galileo Agent}.~Provides intelligent coaching and development recommendations for employees by analyzing performance data, skills assessments, career goals, and learning history to suggest personalized development paths, training programs, and growth opportunities. Accelerates employee development and career progression.\par

\subsubsection*{II-I. Employee Onboarding Coordinator}
\noindent\hangindent=2em\hangafter=1\makebox[2em][l]{\textbf{109.}}\textbf{Onboarding Companion}.~Empowers new hires with real-time support and clarity throughout their onboarding journey, accelerating time-to-value and improving overall employee confidence. Provides contextual guidance, answers questions, and ensures smooth integration into the organization.\par
\noindent\hangindent=2em\hangafter=1\makebox[2em][l]{\textbf{110.}}\textbf{Task Agent}.~Streamlines and automates onboarding tasks so every stakeholder knows exactly what to do and when, improving completion rates and operational efficiency. Manages task assignment, tracking, and escalation across HR, IT, facilities, and managers.\par

\subsubsection*{II-J. Compensation \& Benefits Analyst}
\noindent\hangindent=2em\hangafter=1\makebox[2em][l]{\textbf{111.}}\textbf{Pay Equity Agent}.~Analyzes compensation data to identify and address pay equity gaps across gender, ethnicity, and other protected categories. Provides insights into pay disparities, recommends adjustments, and supports compliance with pay equity regulations while helping organizations achieve fair compensation practices.\par
\noindent\hangindent=2em\hangafter=1\makebox[2em][l]{\textbf{112.}}\textbf{Planning Agent}.~Enables users to change plan or forecast values, or generate additional plan and forecast versions to simulate different outcomes via natural language. Users can accelerate planning cycle time and improve forecast accuracy by creating private simulations, performing direct plan updates, incorporating selected external data, and creating data actions for batch processing and calculations.\par

\subsubsection*{II-K. Payroll Processing Specialist}
\noindent\hangindent=2em\hangafter=1\makebox[2em][l]{\textbf{113.}}\textbf{Payroll Alert Resolution Agent}.~Analyzes payroll errors and finds likely solutions by engaging with other agents to resolve them. Capable of creating validation rules in a payroll control center to prevent recurring errors, achieving 60\% reduction in time to resolve payroll run errors.\par
\noindent\hangindent=2em\hangafter=1\makebox[2em][l]{\textbf{114.}}\textbf{Payroll Anomaly Detection Agent}.~Monitors payroll replication issues in the core HR system, performs root cause analyses, and proposes fixes with human-in-the-loop approval. Achieves 25\% reduction in time to resolve integration errors through proactive detection.\par
\noindent\hangindent=2em\hangafter=1\makebox[2em][l]{\textbf{115.}}\textbf{Payroll Explanation Agent With Benefits}.~Answers questions on pay statements including time management related questions, helping employees understand their compensation details. Reduces time spent on employee pay slip tickets by 50\% through instant, accurate explanations.\par
\noindent\hangindent=2em\hangafter=1\makebox[2em][l]{\textbf{116.}}\textbf{Time Service Agent}.~Capable of creating validation rules in the payroll control center based on natural language input, enabling payroll administrators to define complex business rules without technical expertise. Reduces rule creation time by 70\%.\par

\subsubsection*{II-L. Skills \& Competency Manager}
\noindent\hangindent=2em\hangafter=1\makebox[2em][l]{\textbf{117.}}\textbf{Skills Governance Agent}.~Manages and maintains skills taxonomy and governance by standardizing skill definitions, identifying skill redundancies, suggesting skill updates, and ensuring consistency across the organization. Enables accurate skills tracking, talent management, and workforce planning.\par

\subsubsection*{II-M. HR Service Delivery}
\noindent\hangindent=2em\hangafter=1\makebox[2em][l]{\textbf{118.}}\textbf{Ticket Deflection Agent}.~Instead of directly creating a ESM service ticket, the agent reasons for self-services and guidance in documents, and proposes the dedicated self-services and how to guidance. If no answer is found, create a ticket with the context.\par

\subsubsection*{II-N. Case Handling Specialist}
\noindent\hangindent=2em\hangafter=1\makebox[2em][l]{\textbf{119.}}\textbf{Case Preparation Agent Case Handling Assistant}.~Prepares customer support cases before they reach a human agent. It determines case type and category, extracts provided information into structured fields, detects priority and sentiment, summarizes the customer issue, asks follow-up questions for missing data, routes the case to the appropriate support queue, and enriches the case with historical context.\par
\noindent\hangindent=2em\hangafter=1\makebox[2em][l]{\textbf{120.}}\textbf{Case Processing Agent Case Handling Assistant}.~Operates after a customer case is created. It analyzes the customer issue, reviews available data, predicts the most probable resolution, suggests next-best actions, recommends a resolution plan, drafts an email reply, and automates standard backend actions when appropriate.\par

\subsubsection*{II-O. HR Data Review Specialist}
\noindent\hangindent=2em\hangafter=1\makebox[2em][l]{\textbf{121.}}\textbf{Employee Data Integration Agent AI Assisted CV Summarization Assistent}.~Capable of monitoring issues in the core-HR-to-payroll replication, doing root cause analyses, proposing the fix with human-in-the-loop.\par

\subsection*{III. Procurement}
\subsubsection*{III-A. Strategic Sourcing Specialist}
\noindent\hangindent=2em\hangafter=1\makebox[2em][l]{\textbf{122.}}\textbf{Bid Analysis Agent}.~Compares supplier bids, normalizes responses, and highlights risks and trade-offs across proposals. Automates 80\%+ of bid analysis work by standardizing diverse supplier formats and surfacing key differentiators for decision-makers.\par
\noindent\hangindent=2em\hangafter=1\makebox[2em][l]{\textbf{123.}}\textbf{Executive Summary Agent}.~Produces leadership-ready summaries with key insights, risks, and recommended actions from sourcing events. Distills complex bid analysis, market intelligence, and negotiation outcomes into concise executive briefings.\par
\noindent\hangindent=2em\hangafter=1\makebox[2em][l]{\textbf{124.}}\textbf{Integrated Deep Research Agent}.~Analyzes market signals and supplier performance to identify risk and pricing trends for sourcing decisions. Conducts deep market research combining multiple data sources to inform strategic sourcing.\par
\noindent\hangindent=2em\hangafter=1\makebox[2em][l]{\textbf{125.}}\textbf{Negotiation Intelligence}.~Recommends negotiation moves using past outcomes and market benchmarks. Provides data-driven negotiation strategies based on supplier history, market conditions, and organizational leverage to optimize contract terms.\par
\noindent\hangindent=2em\hangafter=1\makebox[2em][l]{\textbf{126.}}\textbf{RFP Create Agent}.~Automates RFP creation using requirements, historical events, and policy guidance. Generates structured, comprehensive RFP documents that incorporate organizational standards and category-specific requirements.\par
\noindent\hangindent=2em\hangafter=1\makebox[2em][l]{\textbf{127.}}\textbf{RFX Orchestration}.~Automates RFP creation using requirements, historical events, and policy guidance. End-to-end orchestration of the sourcing event process from requirement gathering through supplier responses, ensuring compliance with procurement policies.\par
\noindent\hangindent=2em\hangafter=1\makebox[2em][l]{\textbf{128.}}\textbf{Strategy Recommendation Agent}.~Suggests optimal sourcing strategies using predictive insights and scenario analysis. Evaluates multiple sourcing approaches against organizational goals, market conditions, and risk tolerance to recommend the best path forward.\par
\noindent\hangindent=2em\hangafter=1\makebox[2em][l]{\textbf{129.}}\textbf{Supplier Discovery}.~Analyzes market signals and supplier performance to identify risk and pricing trends, discovering optimal suppliers for sourcing needs. Combines internal spend data with external market intelligence to recommend qualified suppliers.\par

\subsubsection*{III-B. Contract Lifecycle Manager}
\noindent\hangindent=2em\hangafter=1\makebox[2em][l]{\textbf{130.}}\textbf{Compliance Monitoring}.~Continuous monitoring of contract compliance and obligation tracking.\par
\noindent\hangindent=2em\hangafter=1\makebox[2em][l]{\textbf{131.}}\textbf{Contract Authoring}.~AI-assisted contract drafting with clause library and risk assessment.\par
\noindent\hangindent=2em\hangafter=1\makebox[2em][l]{\textbf{132.}}\textbf{Intelligent Contract QA}.~Provides intelligent question-answering capabilities for contract documents by extracting key terms, obligations, dates, and clauses from complex legal agreements. Enables users to quickly find specific information within contracts without manual review, accelerating contract analysis and decision-making.\par
\noindent\hangindent=2em\hangafter=1\makebox[2em][l]{\textbf{133.}}\textbf{Renewal Optimization}.~Proactive identification and optimization of contract renewal opportunities.\par

\subsubsection*{III-C. Invoice Processing Specialist}
\noindent\hangindent=2em\hangafter=1\makebox[2em][l]{\textbf{134.}}\textbf{Invoice Capture Agent}.~Extracts information from invoices sent as PDF or images leveraging multiple AI engines including M/L and LLMs to achieve the highest levels of accuracy. Achieves 90\%+ capture accuracy by combining multiple extraction methods for robust data recognition.\par
\noindent\hangindent=2em\hangafter=1\makebox[2em][l]{\textbf{135.}}\textbf{Invoice Capture LLM Based OCR}.~Extracts information from invoices using advanced LLM-based OCR technology for superior accuracy on complex document layouts. Handles multi-language invoices, varied formats, and poor-quality scans with AI-enhanced recognition.\par
\noindent\hangindent=2em\hangafter=1\makebox[2em][l]{\textbf{136.}}\textbf{Invoice Dispute Management Agent}.~Handles invoice dispute resolution with suppliers via the procurement network AI agent, including invoice rejection, creation of Credit or Debit Memos, incorrect unit price resolution, and other dispute types. Automates communication and resolution tracking.\par
\noindent\hangindent=2em\hangafter=1\makebox[2em][l]{\textbf{137.}}\textbf{Invoice Exception Handling Agent}.~Handles invoice processing exceptions by itself, or coordinates with other agents to provide alternatives for the end-user on how they should resolve the exception based on previous actions. Focused on missing receipts first, reduces manual intervention in exception management.\par
\noindent\hangindent=2em\hangafter=1\makebox[2em][l]{\textbf{138.}}\textbf{Invoice Fraud Prevention Detection Agent}.~Identifies fraudulent patterns not only within a single invoice, but across invoices from a supplier and across suppliers for a customer. Detects multi-transaction anomaly patterns before payment and flags suspicious cases for investigation.\par
\noindent\hangindent=2em\hangafter=1\makebox[2em][l]{\textbf{139.}}\textbf{Invoice Reconciliation Agent 23 Way Match}.~Performs 2/3-way matching and invoice reconciliation processing that evaluates tolerances, limits, previous received invoices, custom fields, and even customer-specific logic to determine the match between invoice and PO line-items.\par
\noindent\hangindent=2em\hangafter=1\makebox[2em][l]{\textbf{140.}}\textbf{Invoice Tax Assistant Agent}.~Helps identify Tax Codes, G/L Accounts, WBS and other codes defined by the buyer related to a supplier invoice. For non-PO invoices, it uses M/L and Tabular AI to propose codes based on previous invoices, reducing manual data entry.\par
\noindent\hangindent=2em\hangafter=1\makebox[2em][l]{\textbf{141.}}\textbf{Invoicing Codification Agent}.~Helps identify Tax Codes, G/L Accounts, WBS and other codes defined by the buyer related to a supplier invoice. For non-PO invoices, uses M/L and Tabular AI to propose codes based on previous invoices for faster processing.\par
\noindent\hangindent=2em\hangafter=1\makebox[2em][l]{\textbf{142.}}\textbf{Invoicing Fraud Prevention Agent}.~Identifies fraudulent patterns not only in a single invoice, but across all invoices from a supplier and all invoices from all suppliers of a customer. Uses AI to detect anomalies and prevent costly fraud incidents before payment.\par
\noindent\hangindent=2em\hangafter=1\makebox[2em][l]{\textbf{143.}}\textbf{Missing Receipts Agent}.~Focuses on resolving invoice exceptions caused by missing goods receipts. Coordinates with other agents and systems to locate receipt information, prompt receipt creation, or recommend alternative resolution paths.\par
\noindent\hangindent=2em\hangafter=1\makebox[2em][l]{\textbf{144.}}\textbf{Missing Reference AI Agents With GenAI}.~Identifies and resolves missing reference data issues in master data management using GenAI capabilities to suggest appropriate values, match similar entities, and recommend corrections. Improves data quality by automatically handling incomplete records and reference gaps in enterprise systems.\par

\subsubsection*{III-D. Travel \& Expense Manager}
\noindent\hangindent=2em\hangafter=1\makebox[2em][l]{\textbf{145.}}\textbf{Receipt Analysis Agent}.~The receipt analysis agent leverages multiple tools, including maps, vendor databases, web searches, and trip itineraries from a travel management platform, to itemize and categorize receipt data for business expense submitters. It reasons over data on the receipt and contextual data outside the receipt, such as map data and business trip itineraries, to accurately create expense entries.\par

\subsection*{IV. Customer Support}
\subsubsection*{IV-A. Customer Support Specialist}
\noindent\hangindent=2em\hangafter=1\makebox[2em][l]{\textbf{146.}}\textbf{Agent Handoff Framework For Support Ticket Creation And Live Agent Support}.~Agent Handoff framework for support ticket creation and/or live agent, via API complete with a detailed conversation summary, when the AI agent is unable to resolve the issue\par
\noindent\hangindent=2em\hangafter=1\makebox[2em][l]{\textbf{147.}}\textbf{Case Classification Agent}.~Analyzes new customer service tickets or cases and classifies them according to business requirements. The classification supports routing, prioritization, and consistent handling of incoming service work.\par
\noindent\hangindent=2em\hangafter=1\makebox[2em][l]{\textbf{148.}}\textbf{Contextual Handoff From Self Service Agents}.~Recommend support options when the AI Agent is unable to answer questions\par
\noindent\hangindent=2em\hangafter=1\makebox[2em][l]{\textbf{149.}}\textbf{Digital Self Service Agent Hand Off For Case Creation}.~The digital service agent facilitates intent identification for ticket creation via conversation while capturing essential information required for hand-off to underlying applications, such as the service management platform or any other application for ticketing of choice by the business. Utilizing the provided hand-off information, these applications employ their own APIs to efficiently create cases, streamlining the process and ensuring accurate resolution.\par
\noindent\hangindent=2em\hangafter=1\makebox[2em][l]{\textbf{150.}}\textbf{Digital Service Agent}.~Responds to customer inquiries in natural language using enterprise application knowledge. The agent handles common questions and service issues, escalates complex cases when needed, and helps support teams focus on interactions requiring human judgment.\par
\noindent\hangindent=2em\hangafter=1\makebox[2em][l]{\textbf{151.}}\textbf{Knowledge Creation Agent}.~Analyzes new tickets or cases in customer service, sales, or service-management platforms and drafts a knowledge-base article that can be linked back to the source case. The agent turns resolved cases into reusable support content.\par
\noindent\hangindent=2em\hangafter=1\makebox[2em][l]{\textbf{152.}}\textbf{New Trigger Options For Custom AI Agents}.~Extend the `trigger' options to provide more flexibility, provide new Agent Tools to broaden the use cases available, and add more output options to increase efficiencies. - Add new Trigger Operation for record update - Add new customer-relationship trigger entities (accounts, leads, contacts)\par
\noindent\hangindent=2em\hangafter=1\makebox[2em][l]{\textbf{153.}}\textbf{QA Agent}.~This agent can autonomously process natural language questions, identify intent, and deliver accurate responses. The agent allows administrators to define particular questions that are answered using existing custom data sources, tickets, or cases.\par
\noindent\hangindent=2em\hangafter=1\makebox[2em][l]{\textbf{154.}}\textbf{Quote Creation Agent}.~Automates quote generation in a customer service platform by monitoring an email integration, extracting relevant request details, and transferring structured information into the quoting workflow.\par

\subsubsection*{IV-B. Commerce \& Shopping Experience}
\noindent\hangindent=2em\hangafter=1\makebox[2em][l]{\textbf{155.}}\textbf{Shopping Agent}.~Supports conversational product discovery in an e-commerce platform. The agent interprets natural-language shopping requests, compares products, uses catalog context, and recommends relevant items for multi-part customer queries.\par
\noindent\hangindent=2em\hangafter=1\makebox[2em][l]{\textbf{156.}}\textbf{Shopping Agent Product Page Context Awareness}.~Adds product-page context to conversational shopping assistance. The agent uses the currently viewed item, related catalog data, and customer questions to compare products, explain compatibility, and recommend alternatives or complements.\par

\subsection*{V. Technology Platform}
\subsubsection*{V-A. IT Operations Specialist}
\noindent\hangindent=2em\hangafter=1\makebox[2em][l]{\textbf{157.}}\textbf{Accounting Accruals Agent}.~Automates journal-entry preparation for accruals by analyzing historical data and accounting policy documents. It calculates proposed accrual amounts and produces pre-populated journal entries for review and confirmation during period-end close.\par

\subsubsection*{V-B. Process Mining \& Transformation}
\noindent\hangindent=2em\hangafter=1\makebox[2em][l]{\textbf{158.}}\textbf{Dashboard Analyzer Agent}.~The Dashboard Analyzer Agent for the process mining platform transforms complex process mining data into clear, actionable insights. It autonomously interprets event logs and KPIs, identifies inefficiencies and deviations, and recommends next best actions--turning static dashboards into intelligent, prescriptive tools. Embedded in the process mining platform, the agent delivers natural-language summaries that explain what's happening, why it matters, and how to improve, enabling faster decisions, reduced analysis time, and continuous process optimization.\par
\noindent\hangindent=2em\hangafter=1\makebox[2em][l]{\textbf{159.}}\textbf{Process Content Recommender Agent}.~Recommends process content for specific transformation questions. The agent reasons over reference and custom process models to answer user questions and provide prioritized recommendations such as value accelerators, KPIs, and process models.\par
\noindent\hangindent=2em\hangafter=1\makebox[2em][l]{\textbf{160.}}\textbf{Screen Guide Agent}.~The Screen Explainer Agent for the process mining platform transforms complex, feature-rich screens into clear, actionable understanding. It dynamically recognizes the screen a user is viewing, explains its purpose, guides users on available actions, and highlights the most relevant data and controls--turning overwhelming dashboards, models, or analyses into intuitive, context-aware experiences. Embedded directly in the process mining platform, the agent delivers natural-language explanations, visual highlights, and role-specific tips that show users what matters most, how to navigate, and which actions to take. By shortening onboarding, improving comprehension, and focusing attention on key insights, the Screen Explainer Agent accelerates enablement, boosts productivity, and empowers users of all experience levels to confidently explore, interpret, and act within the platform.\par
\noindent\hangindent=2em\hangafter=1\makebox[2em][l]{\textbf{161.}}\textbf{Value Case Creation Agent}.~The Value Case Creation Agent helps organizations translate process insights into tangible business value. It automatically identifies inefficiencies, such as rework or delays, and links them to improvement opportunities like automation or standardization. By estimating the potential financial impact using configurable cost and effort parameters, it enables fast, data-driven value case creation without requiring deep analytical expertise. Users receive editable value case drafts with problem summaries, root causes, and expected benefits. These drafts can be validated, refined, and shared across teams. This streamlines collaboration between business and technical stakeholders, accelerates ROI justification, and ensures evidence-based prioritization of transformation initiatives. The agent empowers organizations to confidently move from process findings to financial impact, unlocking measurable value from process improvement efforts.\par
\noindent\hangindent=2em\hangafter=1\makebox[2em][l]{\textbf{162.}}\textbf{Workspace Administration Agent}.~Automates user onboarding and workspace management for a process mining platform. It creates users, assigns roles and access rights, enrolls users in relevant process analyses, and enforces consistent governance for collaborative workspaces.\par

\subsubsection*{V-C. Integration Operations Specialist}
\noindent\hangindent=2em\hangafter=1\makebox[2em][l]{\textbf{163.}}\textbf{Message Processing Error Resolution Agent}.~GenAI-assisted error resolution for failed messages in the message monitor of an integration platform\par

\subsubsection*{V-D. Manufacturing Operations Manager}
\noindent\hangindent=2em\hangafter=1\makebox[2em][l]{\textbf{164.}}\textbf{Production Planning Operations Agent}.~Automates prerequisite checks for releasing production orders, including material availability and capacity availability. It flags material shortages, suggests workarounds such as alternative components or schedule adjustments, and supports supervisor approval before order release.\par

\subsection*{VI. Revenue}
\subsubsection*{VI-A. Demand Generation Strategist}
\noindent\hangindent=2em\hangafter=1\makebox[2em][l]{\textbf{165.}}\textbf{Campaign Agent}.~Designs and orchestrates personalized multi-channel campaigns by combining segmentation insights with predefined playbooks. Continuously monitors campaign performance (opens, clicks, responses) and dynamically optimizes outreach to maximize engagement and conversion.\par
\noindent\hangindent=2em\hangafter=1\makebox[2em][l]{\textbf{166.}}\textbf{Segmentation Agent Engagement Assistant}.~Analyzes behavioral and firmographic signals (such as website activity, engagement patterns, and intent data) to identify high-potential prospects. Produces a refined, prioritized audience list for targeted campaign execution.\par
\noindent\hangindent=2em\hangafter=1\makebox[2em][l]{\textbf{167.}}\textbf{Segmentation Agent Demand Generation Assistant}.~Analyzes behavioral and firmographic signals (such as website activity, engagement patterns, and intent data) to identify high-potential prospects. Produces a refined, prioritized audience list for targeted campaign execution.\par

\subsubsection*{VI-B. Employee Self-Service Portal}
\noindent\hangindent=2em\hangafter=1\makebox[2em][l]{\textbf{168.}}\textbf{Case Preparation Agent Self Service Assistant}.~Prepares customer support cases before they reach a human agent. It determines case type and category, extracts provided information into structured fields, detects priority and sentiment, summarizes the customer issue, asks follow-up questions for missing data, routes the case to the appropriate support queue, and enriches the case with historical context.\par
\noindent\hangindent=2em\hangafter=1\makebox[2em][l]{\textbf{169.}}\textbf{HR Service Ticket Deflection Agent}.~Deflects HR service tickets by providing instant, accurate answers to common HR questions and guiding employees through self-service options before ticket creation. Reduces ticket volume, accelerates resolution times, and improves employee experience in the HR management platform environments.\par
\noindent\hangindent=2em\hangafter=1\makebox[2em][l]{\textbf{170.}}\textbf{HR Utilities Self Service Agent}.~Provides self-service capabilities for utilities and HR employees to access information, update personal data, request services, and resolve common issues without requiring agent assistance. Empowers employees with instant access to HR policies, benefits information, and transactional capabilities.\par

\subsubsection*{VI-C. Merchandising \& Retail Operations}
\noindent\hangindent=2em\hangafter=1\makebox[2em][l]{\textbf{171.}}\textbf{Catalog Assortment Agent}.~Optimizes retail catalog and product assortment strategies by analyzing sales data, customer preferences, market trends, and inventory performance to recommend which products to carry, discontinue, or promote. Helps retailers maximize category performance and meet customer demand while optimizing inventory investments.\par
\noindent\hangindent=2em\hangafter=1\makebox[2em][l]{\textbf{172.}}\textbf{Demand Replenishment Agent}.~Forecasts demand and automates replenishment decisions for retail operations by analyzing historical sales, seasonal patterns, promotional impacts, and inventory levels to ensure optimal stock availability while minimizing excess inventory and stockouts.\par

\subsubsection*{VI-D. Service Operations Manager}
\noindent\hangindent=2em\hangafter=1\makebox[2em][l]{\textbf{173.}}\textbf{Compliance Agent Service Manager Assistant}.~Monitors regulations, tracks individual compliance, auto-enrolls users, manages prerequisites, sends proactive alerts.\par
\noindent\hangindent=2em\hangafter=1\makebox[2em][l]{\textbf{174.}}\textbf{Compliance Agent Service Manager Assistant 2}.~Monitors regulations, tracks individual compliance, auto-enrolls users, manages prerequisites, sends proactive alerts.\par

\subsubsection*{VI-E. Sales Operations Manager}
\noindent\hangindent=2em\hangafter=1\makebox[2em][l]{\textbf{175.}}\textbf{Deal Manager Agent}.~Manages sales deal lifecycle from opportunity creation through negotiation, approvals, and closing by tracking deal status, coordinating stakeholders, managing approvals, and providing insights on deal health and win probability. Accelerates deal velocity and improves sales effectiveness.\par
\noindent\hangindent=2em\hangafter=1\makebox[2em][l]{\textbf{176.}}\textbf{Meeting Assistant Agent}.~Assists sales teams with meeting preparation, scheduling, note-taking, and follow-up actions by extracting key discussion points, capturing action items, identifying next steps, and updating customer relationship management records. Enhances meeting productivity and ensures critical information is captured and actioned.\par
\noindent\hangindent=2em\hangafter=1\makebox[2em][l]{\textbf{177.}}\textbf{Tender Analysis Agent}.~Tender Analysis Agent analyzes dense tender and RFQ documents to automate extraction of critical product requirements, flags risks and policy gaps for quick assessment, and suggests optimized configurations. This reduces manual effort, accelerates sales cycles, improves proposal accuracy, and uncovers new opportunities for cross-sell and up-sell.\par

\subsubsection*{VI-F. Retail Operations Coordinator}
\noindent\hangindent=2em\hangafter=1\makebox[2em][l]{\textbf{178.}}\textbf{Demand Forecasting Agent}.~Provides accurate demand forecasts across multiple time horizons by leveraging advanced analytics, machine learning, and external factors (market trends, weather, events) to predict future demand. Enables better planning, inventory optimization, and supply chain decisions.\par

\subsubsection*{VI-G. Skill \& Qualification Assessor}
\noindent\hangindent=2em\hangafter=1\makebox[2em][l]{\textbf{179.}}\textbf{Engagement Agent}.~Initiates and manages contextual customer outreach and responses to ensure zero delay in lead response. Creates personalized engagement sequences based on lead profile, intent signals, and optimal communication timing.\par
\noindent\hangindent=2em\hangafter=1\makebox[2em][l]{\textbf{180.}}\textbf{Lead Conversion Agent}.~Detects buying intent and converts qualified leads into accounts, contacts, and opportunities. Monitors behavioral signals to identify conversion-ready leads and automatically creates the appropriate customer relationship management records to accelerate deal progression.\par
\noindent\hangindent=2em\hangafter=1\makebox[2em][l]{\textbf{181.}}\textbf{Lead Flow Agent}.~Orchestrates lead enrichment, pre-qualification, and readiness assessment to determine the next best action for each incoming lead. Combines data enrichment with scoring algorithms to ensure only qualified leads progress through the funnel, reducing manual effort by 70\%.\par
\noindent\hangindent=2em\hangafter=1\makebox[2em][l]{\textbf{182.}}\textbf{Lead Routing}.~Assigns qualified leads to the right seller based on capacity, rules, and fit to achieve 90\% reduction in lead reassignment. Uses intelligent matching to pair leads with the most appropriate sales representative for maximum conversion probability.\par

\subsubsection*{VI-H. Period-End Close Coordinator}
\noindent\hangindent=2em\hangafter=1\makebox[2em][l]{\textbf{183.}}\textbf{Order Assistant Agent}.~Provides intelligent assistance for order management by helping users check order status, resolve order issues, expedite shipments, modify orders, and answer order-related queries. Reduces manual effort in order operations and improves customer service response times.\par

\subsubsection*{VI-I. Commerce \& Shopping Experience}
\noindent\hangindent=2em\hangafter=1\makebox[2em][l]{\textbf{184.}}\textbf{Pricing Promotion Loyalty Agent}.~Optimizes retail pricing, promotions, and loyalty programs by analyzing customer behavior, competitive pricing, inventory levels, and market trends to recommend dynamic pricing strategies, targeted promotions, and personalized loyalty offers that maximize revenue and customer engagement.\par

\subsubsection*{VI-J. Order Lifecycle Manager}
\noindent\hangindent=2em\hangafter=1\makebox[2em][l]{\textbf{185.}}\textbf{Returns Claims Agent}.~Streamlines retail returns and claims processing by automating return authorization, validating return eligibility, processing refunds or exchanges, and managing warranty claims. Reduces processing time, improves customer satisfaction, and minimizes fraud while handling complex return scenarios.\par

\subsubsection*{VI-K. Debt Recovery Specialist}
\noindent\hangindent=2em\hangafter=1\makebox[2em][l]{\textbf{186.}}\textbf{Subscription Lifecycle Agent}.~Manages the complete subscription lifecycle including provisioning, renewals, upgrades, downgrades, and cancellations by automating workflows, tracking subscription health, identifying renewal opportunities, and proactively addressing churn risks. Optimizes subscription revenue and customer retention.\par

\subsubsection*{VI-L. Customer Self-Service Specialist}
\noindent\hangindent=2em\hangafter=1\makebox[2em][l]{\textbf{187.}}\textbf{Utilities Customer Self Service Agent}.~The utilities customer self-service agent delivers fast, personalized answers in multiple languages and integrates with enterprise ERP solutions for utilities. It provides tailored customer experiences with an understanding of customer context, including contracts, tariffs, product details, consumption data, and up-selling opportunities. The agent helps utility organizations address challenges stemming from market deregulation and the rise of prosumers, which increase the volume and complexity of customer interactions.\par

\subsubsection*{VI-M. Collections Specialist}
\noindent\hangindent=2em\hangafter=1\makebox[2em][l]{\textbf{188.}}\textbf{Case Preparation Agent Collection Assistant}.~Prepares customer support cases before they reach a human agent. It determines case type and category, extracts provided information into structured fields, detects priority and sentiment, summarizes the customer issue, asks follow-up questions for missing data, routes the case to the appropriate support queue, and enriches the case with historical context.\par

\subsection*{VII. Supply Chain}
\subsubsection*{VII-A. Asset Lifecycle Manager}
\noindent\hangindent=2em\hangafter=1\makebox[2em][l]{\textbf{189.}}\textbf{Alert Processing Agent}.~Processes critical asset condition alerts by triggering enriched notifications with data-backed recommended actions. When a critical asset condition is detected, the agent immediately analyzes the situation and provides maintenance teams with actionable intelligence for rapid response.\par
\noindent\hangindent=2em\hangafter=1\makebox[2em][l]{\textbf{190.}}\textbf{Asset Health Agent}.~Provides comprehensive health checks for equipment based on thresholds and anomalies detected through sensor data and operational parameters. Continuously monitors asset condition to predict failures before they occur and recommend preventive actions.\par
\noindent\hangindent=2em\hangafter=1\makebox[2em][l]{\textbf{191.}}\textbf{Maintenance Request Agent}.~Automatically triggers emergency maintenance notifications and creates maintenance orders in an enterprise ERP system when equipment thresholds are met. Ensures zero delay in emergency processing by instantly creating and routing work orders.\par
\noindent\hangindent=2em\hangafter=1\makebox[2em][l]{\textbf{192.}}\textbf{Serv Dispatcher Agent}.~Provides intelligent dispatch by assigning the right technician to speed up resolution of maintenance and service issues. Considers technician skills, location, availability, and task urgency to optimize assignment and minimize response time.\par
\noindent\hangindent=2em\hangafter=1\makebox[2em][l]{\textbf{193.}}\textbf{Technician Agent Mobile}.~Provides a mobile-ready technician briefing that unifies all relevant data to enable fast, confident on-site fixes. Consolidates equipment history, similar past repairs, spare parts availability, and safety information into a single actionable briefing.\par

\subsubsection*{VII-B. Product Design \& Engineering}
\noindent\hangindent=2em\hangafter=1\makebox[2em][l]{\textbf{194.}}\textbf{Recipe Formulation Agent}.~The Recipe Formulation Agent finds, reuses, and adapts existing recipes and ingredients based on structured specification data -- reducing redundant creation and accelerating formulation cycles.\par

\subsubsection*{VII-C. Manufacturing Operations Manager}
\noindent\hangindent=2em\hangafter=1\makebox[2em][l]{\textbf{195.}}\textbf{Shop Floor Supervisor Agent}.~This agent helps production supervisors adjust schedules and reallocate orders to address the potential disruptions on the shop floor, ensuring a seamless and efficient flow of goods. Production supervisors can achieve overall operational efficiency, minimizing downtime by allowing the shop floor to adapt quickly to changes and maintain optimal productivity.\par

\subsubsection*{VII-D. Procurement Operations Specialist}
\noindent\hangindent=2em\hangafter=1\makebox[2em][l]{\textbf{196.}}\textbf{Supplier Discovery Agent}.~Identifies best-fit suppliers and engagement paths to increase margins and resilience.\par
\noindent\hangindent=2em\hangafter=1\makebox[2em][l]{\textbf{197.}}\textbf{Supplier Onboarding Agent}.~Guides supplier onboarding for a procurement network by helping users upload vendor data, plan onboarding waves, match suppliers already present in the network, send invitations, and monitor post-invitation exceptions that require buyer input.\par

\subsection*{VIII. Cross-domain utilities}
\subsubsection*{VIII-A. General utility agents}
\noindent\hangindent=2em\hangafter=1\makebox[2em][l]{\textbf{198.}}\textbf{Create PPT}.~Create PPT\par
\noindent\hangindent=2em\hangafter=1\makebox[2em][l]{\textbf{199.}}\textbf{Create Word Doc}.~Create Word Doc\par
\noindent\hangindent=2em\hangafter=1\makebox[2em][l]{\textbf{200.}}\textbf{Web Search}.~Web Search\par

\endgroup
\FloatBarrier
\end{document}